\documentclass[preprint,12pt]{article}

\usepackage{graphicx}
\usepackage{amssymb}

\usepackage{multirow}
\usepackage[table,xcdraw]{xcolor}
\usepackage{amsmath}

\title{A study on a Q-Learning algorithm application to a manufacturing assembly problem

	\thanks{\textit{\underline{Citation}}: 
	\textbf{Neves, M., Vieira, M., Neto, P. (2021). A study on a Q-learning algorithm application to a manufacturing assembly problem. Journal of Manufacturing Systems, 59, 426–440. https://doi.org/10.1016/j.jmsy.2021.02.014 }} 

}

\author{
	Miguel Neves, Miguel Vieira, Pedro Neto \\
	University of Coimbra \\
	Coimbra\\
}

\begin{document}
	\maketitle

\begin{abstract}
The development of machine learning algorithms has been gathering relevance to address the increasing modelling complexity of manufacturing decision-making problems. Reinforcement learning is a methodology with great potential due to the reduced need for previous training data, i.e., the system learns along time with actual operation. This study focuses on the implementation of a reinforcement learning algorithm in an assembly problem of a given object, aiming to identify the effectiveness of the proposed approach in the optimisation of the assembly process time. A model-free Q-Learning algorithm is applied, considering the learning of a matrix of Q-values (Q-table) from the successive interactions with the environment to suggest an assembly sequence solution. This implementation explores three scenarios with increasing complexity so that the impact of the Q-Learning\textsc{\char13}s parameters and rewards is assessed to improve the reinforcement learning agent performance. The optimisation approach achieved very promising results by learning the optimal assembly sequence 98.3\% of the times.
\end{abstract}



\section{Introduction}
\label{intro}
Alongside the advent of product customization, industrial manufacturing processes are increasingly more complex, required to be highly flexible, resourceful, and efficient. Their related decision-support problems have been explored in literature with the development of traditional optimisation mathematical models (e.g. LP, IP, MILP or MINLP) and heuristic approaches \cite{Ref1}, and where machine learning algorithms have been gathering particular relevance. However, most machine learning approaches rely on trial and error interactions that requires available data (e.g. supervised learning), which is often unavailable. A current approach to such optimisation problems is based on reinforcement learning (RL).

\begin{figure}[]
  \centering\includegraphics[width=0.7\textwidth]{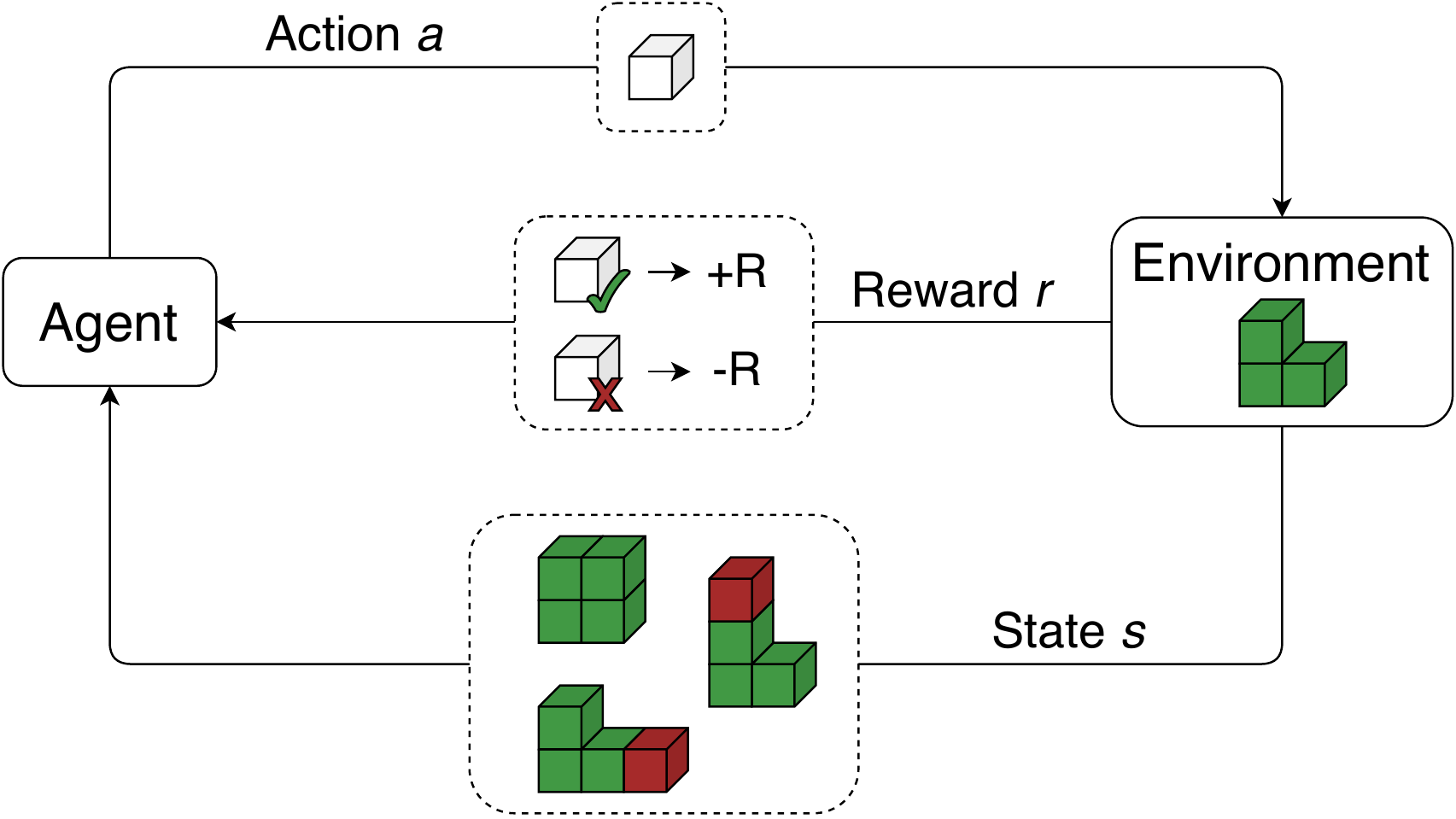}
\caption{Agent and environment interaction.}
\label{fig:1}       
\end{figure}

RL is a machine learning paradigm originally inspired by the way biological systems learn \cite{Ref2}, where an agent (e.g. a human, a robot, a vehicle) interacts with the environment by taking actions, Fig.~\ref{fig:1}. As a consequence of the action taken, the environment defines states and rewards.  A state is essentially a description of the situation of the environment and the rewards are an abstract concept that describes the feedback by which the success or failure of the agent is measured. This kind of learning is gathering relevance in problems that require decision making, by dynamically exploring a solution that maximises the total rewards. 

The main advantages of RL compared to other machine learning approaches are the fewer data available in the learning phase, which are the current states and expected rewards in a continuous learning process, fuelled by the interaction between the agent and the environment. Given its characteristics, RL methods are used in complex problems where there appears to be no obvious or easily programmable solution, such as game playing, robotics, control problems or operational research, and in problems where there is not enough labelled data such as anomaly detection problems. Since most of programming tasks are tedious and require years of expertise, RL algorithms can be applied to replace it with an intuitive process comprehensible even by an unskilled user.

\subsection{Reinforcement learning applications}
\label{sec:intro_1}
One of the most successful applications of RL in recent years was the development of the Go playing program known as AlphaGo. This program achieved a 99.8\% winning rate against other Go programs and defeated the European Go champion by 5 games to 0 \cite{Ref3}. This program was trained by supervised learning from human expert moves and by reinforcement learning from self-play. This application was further improved (AlphaGo Lee) and was capable of defeating the 9 dan player Lee Sedol, winner of 18 international titles, 4 games out of 5 \cite{Ref4}.

Also, in the field of game playing, RL was employed in a set of 49 Atari games by developing the DQN algorithm. This new algorithm was able to outperform the best RL methods at the time in 43 out of the set of 49 games. Furthermore, DQN performed at a level comparable to a professional player, achieving more than 75\% of the human score in 29 out of the set of 49 games tested \cite{Ref5}.

In recent years, the research on the applicability of RL is also increasing in the fields of decision-making and system control problems. By applying Q-Learning in a stock optimisation problem were achieved results up to 25\% better when compared to traditional stock management algorithms, \cite{Ref6}. Wang and Usher studied the implementation of the Q-Learning algorithm for the usage of job agents when establishing routing decisions in a job shop environment, \cite{Ref7}. The authors discussed the effects of the Q-Learning application with the guidelines for future applications and recommendations for factor settings.  Also, in dynamic job shop scheduling problem (DJSS) \cite{Ref8} proposed the usage of RL with a Q-factor algorithm to improve the scheduling method\textsc{\char13}s performance while considering random job arrivals and machine breakdowns. In a simulated environment, this proposed method achieved high performances. In automotive paint shops, colour changeovers between consecutive production orders are a source of costs due to the process of cleaning painting robots. In order to minimize these costs, production orders can be re-sequenced and grouped in identical colours as a batch. Leng et. al. propose the usage of the Deep Q-Network algorithm to solve this Colour-batching Re-sequencing Problem \cite{Ref9}. Huang et al. proved, through a simulation study, the effectiveness of the usage of Q-Learning in a maintenance problem where random failures of machines are highly disruptive \cite{Ref10}. The usage of reinforcement learning was also proposed to tackle the profit optimization problem of a single product production affected by deterioration failures, requiring both maintenance and repairs \cite{Ref11}. The performance in manufacturing work cells that utilise gantries to load and unload the materials and parts needed is highly dependent on the gantry movements in real operation. Ou et al. formulated the gantry scheduling problem as a RL problem through the usage of Q-Learning and demonstrated, from the simulation results, the capability of effectively reducing system production losses in real-time operations, \cite{Ref12}. In a later article, five different reward functions were devised based on different assumptions of the system. It was shown that the policy derived from systematic analyses of production loss significantly outperformed the other policies. Therefore, it was concluded that the level of understanding of the system and how this understanding is transmitted to the reward function greatly impacts the learning model\textsc{\char13}s success \cite{Ref13}. In the areas of production manufacturing \cite{Ref14} proposed the usage of RL to improve assembly efficiency by a dual-arm robot and achieved higher performance when compared with other methods. In recent years, personalized production has emerged due to the increasing customer demand for more personalized products. This type of production, when compared with traditional production, has more uncertainty and variability. Such problems can be tackled by the usage of multi agent systems and reinforcement learning in a smart manufacturing environment. This approach was shown to be competitive in a dynamic environment \cite{Ref15}. The automation of shoemaking production lines is extremely challenging also due to the versatility of manual product customization. To tackle this problem a cyber-physical system artificial intelligence architecture was devised for the complete manufacturing of soft fabric shoe tongues by using Deep-Q reinforcement learning as a means of achieving better control over the manufacturing process and convolutional and long short-term memory artificial neural network to enhance action speed \cite{Ref16}.

In a collaborative assembly process context, \cite{Ref17} proposed an approach based on Interactive Reinforcement Learning to reduce the programming effort required by an expert. The learning approach is made in two steps. The first one consists of modelling simple tasks that constitute the assembly process, using task-based formalism. These modelled simple tasks are then used by the robotic system by proposing to the user a set of possible actions at each step of the assembly process via a graphic user interface (GUI). After the user selects the action, the robot performs it, progressing the assembly process while learning the assembly order. The framework also allows different users to teach different assembly processes to the robot. This proposed approach is based on Q-Learning and IRL and was successfully applied in a UR10 robot in an assembly process comprising tasks such as, picking, holding, mounting and receiving objects. Human-robot interactions have become abundant due to the increase of autonomous robotic operators. The consequence of such interaction is the introduction of a source of uncertainty and unpredictability from the human operator. Oliff et. al. presents a methodology for the implementation of a RL-based intelligent agent which allows a change in the robotic operator\textsc{\char13}s behavioural policy in response to performance variation in the human operators \cite{Ref18}. Lastly, there has been an increase in the research of using digital twins in smart manufacturing environments. Xia et. al. developed a control methodology named Digital Engine capable of acquiring process knowledge, scheduling manufacturing tasks, identifying optimal actions and demonstrating control robustness \cite{Ref19}. Hu et. al. proposed a new graphic convolution layer named Petri-net convolution layer (PNC). When utilizing DQN with a PNC network better solutions are obtained for dynamic scheduling problems \cite{Ref20}.

\subsection{Main challenges and objectives}
\label{sec:intro_2}
Even though many advances were made in recent years, some major challenges where noted by \cite{Ref17} when applying RL algorithms. The first challenge arises in the learning phase where the agent must do a trade-off between doing actions known to be effective (exploitation) and actions not yet explored (exploration), which is known as the exploration-exploitation dilemma \cite{Ref2}. Another common challenge is the “curse of dimensionality” \cite{Ref18}, where to ensure global optimality, data must be collected throughout the entire state-space, often infeasible in high-dimensional state-action spaces. The restrictions of expensive hardware, component\textsc{\char13}s wear with economical and logistical consequences in the agent interactions with the physical world are known as the “curse of real-world samples”. To reduce real-world interactions, a model can be used as a simulation system and the knowledge transferred to a real scenario. However, creating accurate models is very challenging and sometimes even impracticable. Small errors due to under-modelling can accumulate and make the simulation diverge from the real behaviour, which is known as “curse of under-modelling and model uncertainty”. Finally, the desired behaviour in RL is often specified by the reward function, which is frequently easier than defining the behaviour itself, however, in practice, in some problems it may be astonishingly difficult. This RL challenge is often known as “curse of goal specification”.

In this RL application, our approach aims to study the exploration-exploitation dilemma and the curse of goal specification, in which the type of agent properties (e.g. learning rate or reward signal) can influence the results obtained, here tackled by the assessment of the algorithm\textsc{\char13}s learning parameters in the solution\textsc{\char13}s outcome. In addition, this intends to support work planning with the demonstration of the RL method application regarding a feasible and efficient assembly sequence of a product. Therefore, the problem application resumes the optimisation of an assembly problem for a given object considering a time efficient solution.

The remainder of the paper is structured as follows: in Sect.~\ref{sec:2}, the problem formulation and modelling theory of the Q-learning algorithm is outlined; in Sect.~\ref{sec:3}, the case study is presented with the discussion of the results implementation for a set of assembly scenarios; and finalising, in Sect.~\ref{sec:4}, with the conclusions and future work.

\section{Model formulation}
\label{sec:2}

\subsection{Problem description}
\label{sec:2_1}
Due to the market diversification of on-demand product attributes, current production systems are being required to deal with assembly flexibility, in which one or multiple resources is in charge of all steps of product customisation. The tasks sequencing represents one of the major components directly related to the efficiency of the assembly process. However, the assembly of a product is constrained by the product design and quality considerations, which often derives many possibilities in assembly sequence steps.

The problem considers a representation of an assembly job of a complex product containing different parts and tools, decomposed in a number of tasks which can be assembled in different sequences by an assigned resource. Along with the achievement of a feasible assembly process, improving the time efficiency is nontrivial due to the complexity of sequences\textsc{\char13} immediate/general time dependence of previous completed tasks. In order to cope with these issues, the goal of this work is to assess the effectiveness of an RL agent algorithm in the optimisation of task sequences based on real-time system states.

\subsection{Mathematical approach}
\label{sec:2_2}
To provide a baseline comparison, the problem is formulated as a benchmark mathematical optimisation MILP model, given the following mathematical notations:

\begin{table}[h]
\begin{tabular}{p{0.8cm} p{12cm}}
\multicolumn{2}{l}{\textbf{Sets:}}                                                    \\
\textit{i}, \textit{i\textsc{\char13}}, \textit{i\textsc{\char13}\textsc{\char13}} & Tasks                                                                    \\
\textit{k}, \textit{k\textsc{\char13}}      & Sequence step                                                            \\
\multicolumn{2}{l}{\textbf{Subsets:}}                                                 \\
\textit{\(P_{ii'}\)}       & General precedence tasks feasibility                                     \\
\textit{\(Q_{ii'}\)}       & Immediate forbidden sequence of tasks                                    \\
\multicolumn{2}{l}{\textbf{Parameters:}}                                              \\
\textit{\(\tau_{ik}\)}      & Average processing time of task \textit{i} to sequence step \textit{k}                     \\
\textit{\(\Delta_{ii\textsc{\char13}}\)}   & Processing time variation of task \textit{i'} given the previous completed task \textit{i} \\
\multicolumn{2}{l}{\textbf{Variables:}}                                               \\
\textit{\(Y_{ik}\)}        & Binary variable to assign task \textit{i} to sequence step \textit{k}                      \\
\textit{\(C_{ik}\)}        & Completion time of task \textit{i} at each sequence step \textit{k}                        \\
\textit{\(C_{max}\)}       & Makespan                                                                
\end{tabular}
\end{table}

The problem is formulated as an assignment problem of a set of tasks \(i\in I\) to a set of sequence steps \(k\in K\) based on the concept of a general precedence model, setting in Eq.~\ref{eq:1} the objective function as the makespan minimisation. The variable \(C_{ik}\) defines the completion time of task \textit{i} at each step \textit{k}, given the assignment of task \textit{i} to sequence step \textit{k} set by variable 
\(Y_{ik}\). Considering that the last step \(k=K^{last}\) comprises a feasible assembly combining all possible tasks, the total completion time of the optimal assembly sequence is minimised.

\begin{equation}\label{eq:1}
Minimise\; C_{max}\geq C_{ik}\quad \forall i, k=K^{last}
\end{equation}
\textit{s.t.}
\begin{equation}\label{eq:2}
\Sigma_i Y_{ik} = 1\quad \forall k
\end{equation}
\begin{equation}\label{eq:3}
\Sigma_k Y_{ik} = 1\quad \forall i
\end{equation}
\begin{equation}\label{eq:4}
Y_{ik} + Y_{i\textsc{\char13}k\textsc{\char13}} < 1\quad \forall i, i\textsc{\char13}\in P_{ii\textsc{\char13}}, i\neq i\textsc{\char13},k<k\textsc{\char13},k\neq k\textsc{\char13}
\end{equation}
\begin{equation}\label{eq:5}
\begin{aligned}
    & C_{ik}\geq \tau_{ik}Y{ik} + (C_{i\textsc{\char13}k-1}+\Delta_{i\textsc{\char13}i}Y_{i\textsc{\char13}k-1})\mid_{k>1} + \\
    & (\Sigma_{k\textsc{\char13}>k-1}\Sigma_{i\textsc{\char13}\textsc{\char13}\notin Q_{i\textsc{\char13}\textsc{\char13}i}, i\textsc{\char13}\textsc{\char13} \neq i}\Delta_{i\textsc{\char13}\textsc{\char13}i}Y_{i\textsc{\char13}\textsc{\char13}k\textsc{\char13}})\mid_{k>2} \forall i,i\textsc{\char13}\notin Q_{i\textsc{\char13}i,k}
\end{aligned}
\end{equation}
\begin{equation}\label{eq:6}
C_{ik} \geq 0\quad Y_{ik} \in \{0,1\}
\end{equation}

Regarding the formulation, Eqs.~\ref{eq:2}-\ref{eq:3} considers the allocation constraints to allocate one and only one task to one step, and Eq.~\ref{eq:4} guarantees the general precedence rule of two tasks (under subset \(P_{ii\textsc{\char13}}\)). Eq.~\ref{eq:5} defines the completion time for every task and according to production step, ensuring the timing between two consecutive stages, defined by the allowed assembly sequence given by \(Q_{i\textsc{\char13}i}\). Besides the processing time of each task per stage \(\tau_{ik}\), the model considers the variation of time duration according to previous completed tasks, given by \(\Delta_{ii\textsc{\char13}}\). Finally, Eq. ~\ref{eq:6} reassures the non-negativity and integrality of variables.

\subsection{RL modelling theory}
\label{sec:2_3}
Given the anatomy of RL algorithms, they can be classified into three categories of methods which are value-function methods, policy search methods and actor-critic methods. The value-function methods, also known as critic-only methods, are based on the idea of initially discovering the optimal value function by fitting a value-function or a Q-function and then deriving the optimal policy from this. On the other hand, the policy search methods, also known as actor-only methods, search directly in the policy space by summing the rewards of sample trajectories, which is only possible if the search space is restricted. In the particular case of policy search algorithms, known as policy gradient methods, one step of gradient ascent is applied to the expected reward objective. Lastly, the actor-critic methods are a combination of both where the critic\textsc{\char13}s function is to monitor the agent\textsc{\char13}s performance by fitting a value function or a Q-function to determine when the policy must be changed by the actor.  Moreover, the models can also be divided into model-free algorithms, which do not use models of the environment and are explicitly trial-and-error learners, and model-based algorithms, which use the model for planning or policy improvement.

RL algorithms are commonly formalised as Markov decision processes, where the agent observes the current state and decides the next suitable action. The reason for this formalism is the increased difficulty of computation by considering all the states and actions taken from the initial state. Using MDPs, the system only needs to keep track of the last state and action. However, it is important to understand that this Markov assumption leads to the loss of data, which in some situations might be relevant since rewards may be infrequent and delayed.

\begin{figure}[]
  \centering\includegraphics[width=0.8\textwidth]{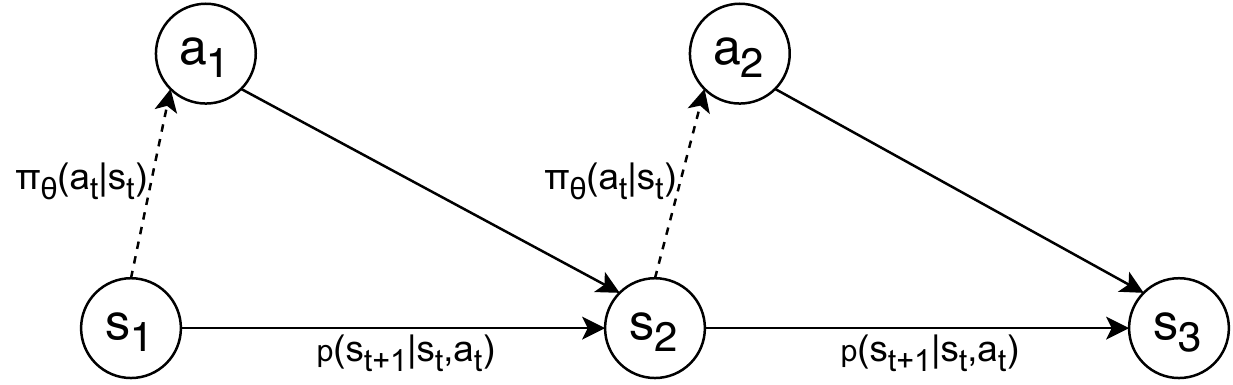}
\caption{Markov decision process (MDP).}
\label{fig:2}       
\end{figure}

To define a Markov decision process, \(M=\{S,A,T,r\}\), it is required to define a state space \textit{S}, an action space \textit{A}, a transition operator \textit{T} and a reward function \textit{r}. The state space is a set of valid states \textit{s} the system can occupy, \(s\in S\). The action space is the set of possible actions \textit{a} the agent can take such that \(a\in A\). In Markov decision processes the transition probabilities are not only conditioned on the previous state but also on the previous action, \(p(s_{t+1}\mid s_t,a_t)\). Since the policy is the mapping of the states to actions, \(\pi_\theta(a_t\mid s_t)\), a graphical representation of a Markov decision process can be observed in Fig.~\ref{fig:2}.

The learning agent must evaluate whether if it is being successful in the task. An agent can discern good from bad events based on the reward signal, which is analogous to the way the humans learn when experiencing pain or pleasure. This is the primary source of improvement of the policy since if the action selected in a certain state returns a low reward, then the policy may be changed so that, when faced by the same exact state, the policy selects a different and more rewarding action. However, the agent\textsc{\char13}s goal is to maximise the accumulated reward over time through the actions chosen. However, an action with a high immediate reward might not be the optimal choice since it may lead to a lower accumulated reward over the future. To tackle this issue, there are two important concepts, the value function and the quality function, usually known as Q-function.
Given a policy, the value function is defined as the total expected reward from a given state \(s_t\):

\begin{equation}\label{eq:7}
V^\pi(s_t) = \sum_{t'=t}^TE_{\pi_\theta}[r(s_{t\textsc{\char13}},a_{t\textsc{\char13}})\mid s_t]
\end{equation}

The Q-function is, on the other hand, the total expected reward from taking the pair action \(a_t\) in the state \(s_t\):

\begin{equation}\label{eq:8}
Q^\pi(s_t,a_t) = \sum_{t'=t}^TE_{\pi_\theta}[r(s_{t\textsc{\char13}},a_{t\textsc{\char13}})\mid s_t,a_t]
\end{equation}

Actions must therefore be selected based on value judgements because the agent\textsc{\char13}s goal is to maximise the accumulated reward over time. Unfortunately, while rewards are given directly by the environment, values must be estimated multiple times within the sequence of observations.

\subsubsection{Q-Learning}
\label{sec:2_3_1}
In order to implement the algorithm, it is necessary to understand the Q-Learning parameters used, such as how the Q-table is updated. The value iteration update is done at each step through the Bellman Equation, which consists on the weighted average of the old Q-value and the new information obtained, where \(\alpha\) corresponds to the learning rate, \(\gamma\) to the discount factor, \(r_t\) to the received reward when moving from state \(s_t\) to \(s_{t+1}\),  \(Q^{new}(s_t,a_t)\) to the new Q-value of the state \(s_t\) and action \(a_t\), \(Q(s_t,a_t)\) to the old Q-value of the state \(s_t\) and action \(a_t\) and \(\max\limits_aQ(s_{t+1},a)\) to the estimate of the optimal future Q-value:

\begin{equation}\label{eq:9}
\begin{aligned}
Q^{new}(s_t&,a_t) \leftarrow Q(s_t,a_t) \:+ \alpha \times [r_t + \gamma \times \max_aQ(s_{t+1},a)-Q(s_t,a_t)]
\end{aligned}
\end{equation}

The learning rate parameter has values between 0 and 1 and influences the extent on how the new information changes the current state value, which means that a lower learning rate leads to a longer learning time. However, it is important to note that a higher learning rate may lead to suboptimal results or even divergence in non-deterministic scenarios. The discount factor determines the importance of future rewards, so the lower its value the less meaningful are the future rewards. If the discount factor has the value 0, only the current reward is considered.
The selection of the action is made using an epsilon greedy search, i.e. the agent selects a random action with probability \(\varepsilon\) and otherwise selects the action greedily, with probability \(1-\varepsilon\), by selecting the action with the highest Q-value. The value of epsilon (\(\varepsilon\)) decays based on a decay rate known as epsilon decay. The algorithm can be summarized as follows:

\begin{table}[h]
\label{alg}       
\begin{tabular}{p{0.5cm} p{12cm}}
\hline\noalign{\smallskip}
\multicolumn{2}{l}{\textbf{Algorithm} Q-Learning} \\
\noalign{\smallskip}\hline\noalign{\smallskip}
01: & \(Q(s,a)\) initialised arbitrarily \\
02: & \textbf{For each} episode \(i\) \textbf{do}  \\
03: & \quad state \(s\) initialised \\
04: & \quad \textbf{For each} step \(j\) \textbf{do} \\
05: & \quad \quad action \(a\) chosen from state \(s\) using policy derived by Q (e.g. \(\varepsilon\)-greedy) \\
06: & \quad \quad action \(a\) taken \\
07: & \quad \quad	reward \(r\) and state s\textsc{\char13} observed \\
08: & \quad \quad	\(Q(s,a) = Q(s,a) + \alpha \times [r + \gamma \times \max\limits_{a\textsc{\char13}}Q(s\textsc{\char13},a\textsc{\char13}) – Q(s,a)]\) \\
09: & \quad \quad \(s = s\textsc{\char13}\) \\
10: & \quad \textbf{end for} \\
11: & \textbf{end for} \\
\noalign{\smallskip}\hline
\end{tabular}
\end{table}

\section{Experiments and results}
\label{sec:3}

\subsection{Case study}
\label{sec:3_1}
\begin{figure*}
  \centering\includegraphics[width=1\textwidth]{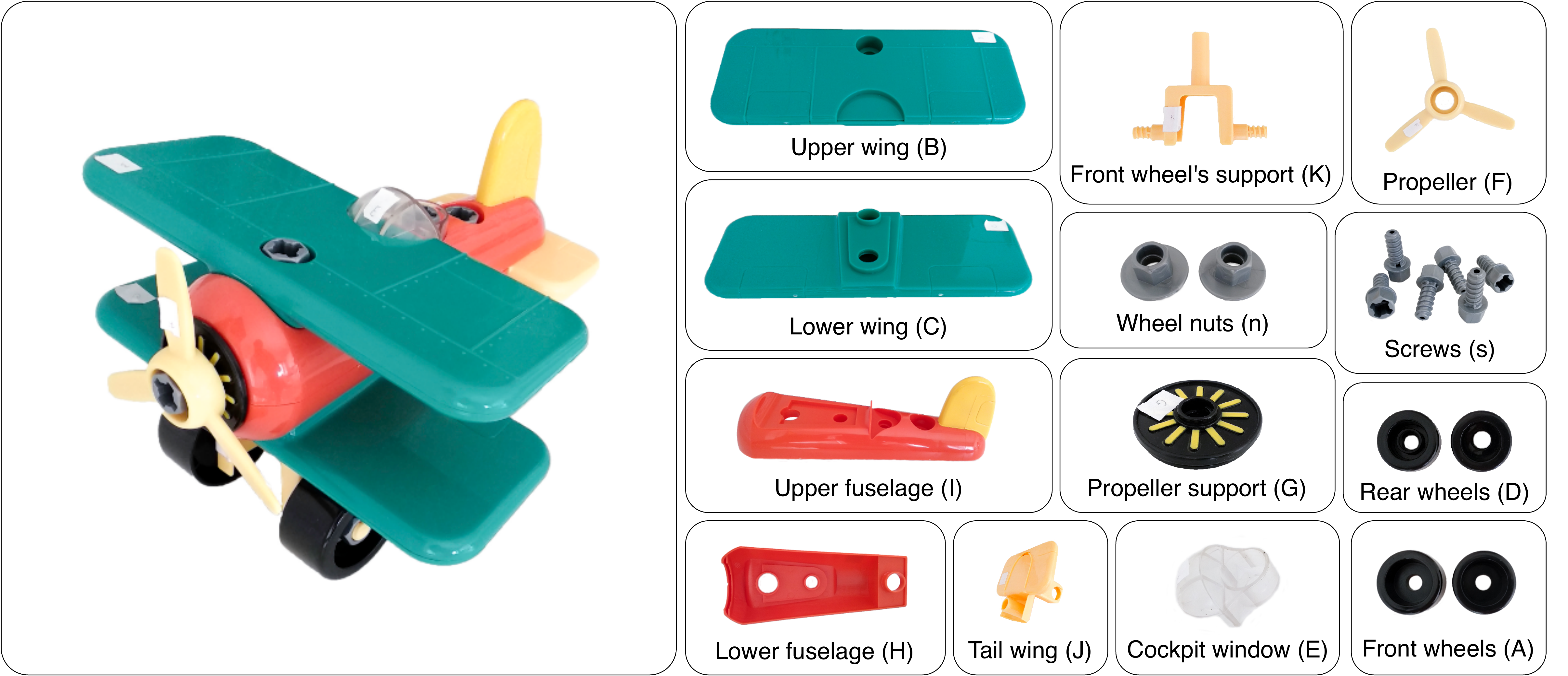}
\caption{Aeroplane from the Yale-CMU-Berkeley Object and Benchmark Dataset.}
\label{fig:3}       
\end{figure*}

The object chosen for the assembly problem is an aeroplane toy from the Yale-CMU-Berkeley Object and Benchmark Dataset \cite{Ref19,Ref20} (Fig.~\ref{fig:3}), which process is studied and optimised through the implementation of a RL methodology.

As follow, the object components and assembly structure are thoroughly analysed. As a preliminary evaluation, the problem solution is derived from the mathematical optimisation model. Then, the assembly problem is formulated using Q-Learning and is implemented in two different scenarios where the agent must learn a feasible assembly sequence, i.e. an assembly sequence that respects all the task precedencies. The scenarios consider the distinct time durations of each tasks incorporated in the decision-making algorithm to foment the learning of an optimised time efficient assembly sequence. The algorithm\textsc{\char13}s learning parameters are individually analysed in order to improve the agent\textsc{\char13}s performance. 

\subsection{Assembly analysis}
\label{sec:3_2}
The aeroplane object is comprised of 9 structural parts and 2 types of fasteners, which are displayed in Table~\ref{tab:1} and Table~\ref{tab:2}.

\begin{table*}[]
\caption{Aeroplane\textsc{\char13}s parts.}
\label{tab:1}       
\begin{tabular}{p{0.7cm}p{8.3cm}p{2.9cm}}
\hline\noalign{\smallskip}
Part & Aeroplane part\textsc{\char13}s description & Number of parts  \\
\noalign{\smallskip}\hline\noalign{\smallskip}
A & Front wheels & 2 \\
B & Upper wing & 1 \\
C & Lower wing & 1 \\
D & Rear wheels & 2 \\
E & Cockpit window & 1 \\
F & Propeller & 1 \\
G & Propeller\textsc{\char13}s support (engine) & 1 \\
H & Lower body of the aeroplane (lower fuselage) & 1 \\
I & Upper body of the aeroplane (upper fuselage) & 1 \\
J & Rear body of the aeroplane (tail wing) & 1 \\
K & Front wheel\textsc{\char13}s support & 1 \\
\noalign{\smallskip}\hline
\end{tabular}
\end{table*}

\begin{table*}[]
\caption{Aeroplane\textsc{\char13}s fasteners.}
\label{tab:2}       
\begin{tabular}{p{4cm}p{4cm}p{4cm}}
\hline\noalign{\smallskip}
Fastener & Fastener\textsc{\char13}s head type & Number of fasteners  \\
\noalign{\smallskip}\hline\noalign{\smallskip}
n & Wheel nuts & 2 \\
s & Screws & 5 \\
\noalign{\smallskip}\hline
\end{tabular}
\end{table*}

After subdividing the aeroplane into parts and fasteners, the assembly process was subdivided in a total of 8 tasks. In the Table~\ref{tab:3} each task is associated with the corresponding parts and fastener required. It is important to note that some parts can be used in more than one task. For the assembly to be complete, every task must be executed without repetitions, therefore the number of different assembly sequences is \(n!=8!=40320\).

\begin{table*}[]
\caption{Parts and fasteners associated to each task and their respective quantities.}
\label{tab:3}       
\begin{tabular}{p{0.8cm}p{0.5cm}p{0.5cm}p{0.5cm}p{0.5cm}p{0.5cm}p{0.5cm}p{0.5cm}p{0.5cm}p{0.5cm}p{0.5cm}p{0.5cm}p{0.7cm}p{0.7cm}}
\hline\noalign{\smallskip}
Tasks & \multicolumn{11}{l}{Parts} & \multicolumn{2}{l}{Fasteners} \\ \cline{2-14} 
                       & A & B & C & D & E & F & G & H & I & J & K & n             & s             \\ 
                       \hline\noalign{\smallskip}
1                      &   &   &   &   &   &   & 1 & 1 & 1 &   &   &               &               \\
2                      &   &   &   &   &   &   &   &   &   & 1 &   & 1             &               \\
3                      &   &   &   &   &   & 1 &   &   &   &   &   & 1             &               \\
4                      &   &   & 1 &   &   &   &   &   &   &   &   & 1             &               \\
5                      &   & 1 &   &   &   &   &   &   &   &   & 1 & 1             &               \\
6                      &   & 1 &   &   & 1 &   &   &   &   &   &   & 1             &               \\
7                      & 1 &   &   &   &   &   &   &   &   &   & 1 &               & 2             \\
8                      &   &   &   & 1 &   &   &   &   &   & 1 &   & 1             &               \\
\noalign{\smallskip}\hline
\end{tabular}
\end{table*}

However, the feasible number of assembly sequences is lower than the previously calculated one, due to the fact that certain tasks require other tasks to be previously completed. Such precedence sequence dependencies are displayed in the Table~\ref{tab:4}. When taking into account the task dependencies, the feasible number of assembly sequences is 3360, which corresponds to only 8.3\% of all assembly sequences.

\begin{table*}[]
\caption{Precedence task\textsc{\char13}s feasibility, \(P_{ii\textsc{\char13}}\).}
\label{tab:4}       
\begin{tabular}{p{5cm}p{7.5cm}}
\hline\noalign{\smallskip}
Task & Precedence task  \\
\noalign{\smallskip}\hline\noalign{\smallskip}
1 & None \\
2 & 1 \\
3 & 1 \\
4 & 1 \\
5 & 1 and 4 \\
6 & 1 \\
7 & None \\
8 & None \\
\noalign{\smallskip}\hline
\end{tabular}
\end{table*}

\begin{figure}
  \centering\includegraphics[width=0.9\textwidth]{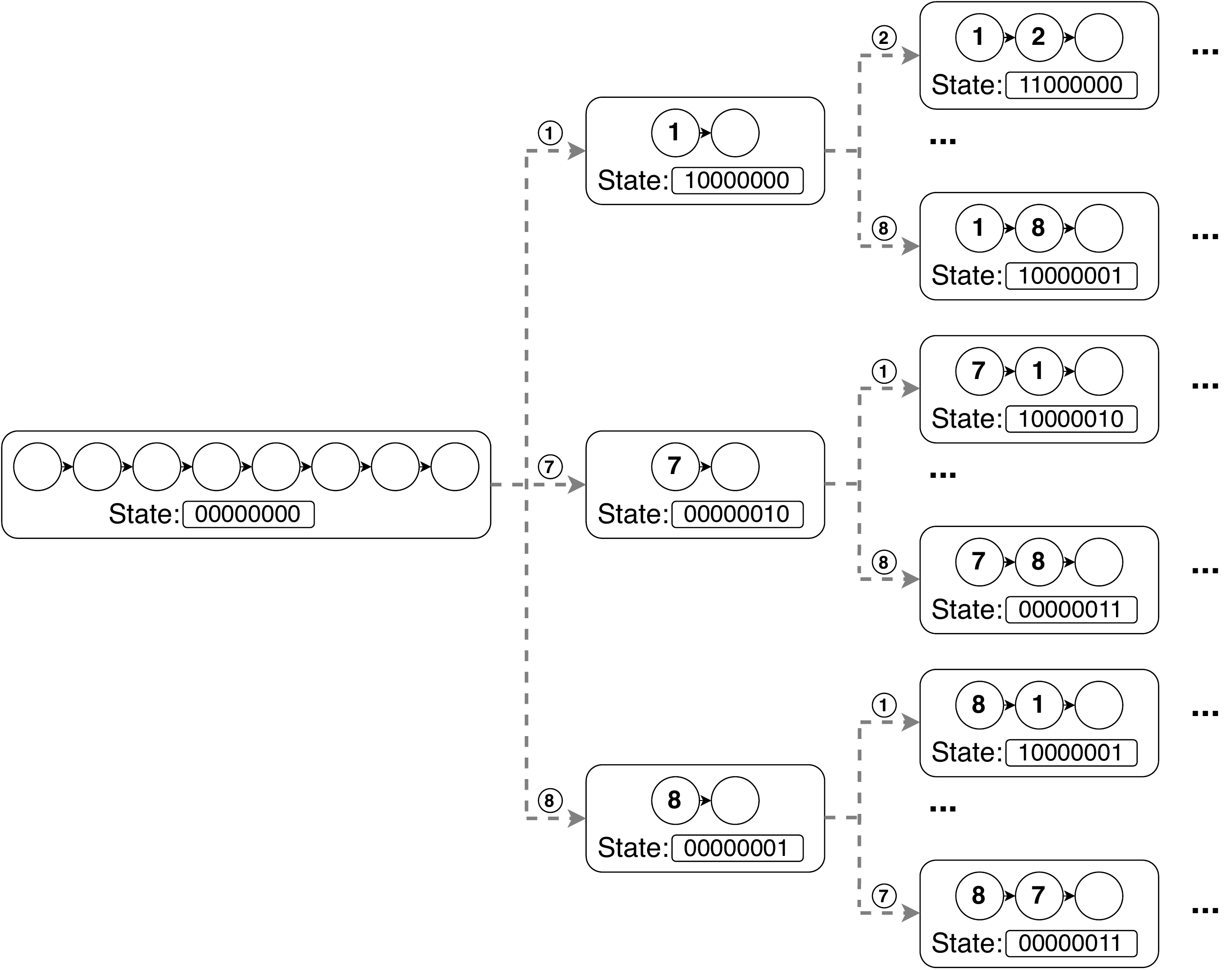}
\caption{MDP\textsc{\char13}s states and actions scheme.}
\label{fig:4}       
\end{figure}

In resume, the assembly process is subdivided in 8 different tasks, or actions, and the assembly can be considered as complete when all the 8 tasks have been executed. Therefore, the MDP\textsc{\char13}s states can be defined by the updated assembly status at each task, where the initial state would correspond to the situation where none of the actions was executed, the final state when all the actions were completed and the aeroplane is assembled, represented by the binary number indicating whether the task has been executed. The digit in the binary number corresponds to the task number. If the task \textit{n} has been executed, 1 is set to the \(n^{th}\) digit in the binary number. Otherwise, the \(n^{th}\) digit is set as 0 (Fig.~\ref{fig:4}). The initial state would then be represented as 00000000 and the final state would correspond to 11111111, which in decimal notation corresponds to 0 and 255 respectively, thus, the number of states is 256. Similarly, due to task precedencies, there are only 100 possible states. 

Finally, for this case, the tasks\textsc{\char13} average time were estimated \(\tau_{ik}\) (Table~\ref{tab:5}) as well as the increase/decrease variances on the average times with respect to the tasks previously done \(\Delta_{ii\textsc{\char13}}\) (Table~\ref{tab:6}). In this table is also shown shaded cells in grey with the immediate forbidden sequences, which add as restrictions to the previous general precedence.

\begin{table*}[]
\caption{Task\textsc{\char13}s average time, \(\tau_{ik}\).}
\label{tab:5}       
\begin{tabular}{p{2.9cm}p{0.8cm}p{0.8cm}p{0.8cm}p{0.8cm}p{0.8cm}p{0.8cm}p{0.8cm}p{0.8cm}}
\hline\noalign{\smallskip}
Task & 1 & 2 & 3 & 4 & 5 & 6 & 7 & 8 \\
\noalign{\smallskip}\hline\noalign{\smallskip}
Average time [time units, t.u.] & 10 & 7 & 8 & 6 & 12 & 8 & 11 & 9 \\
\noalign{\smallskip}\hline
\end{tabular}
\end{table*}

\begin{table}[]
\caption{Tasks\textsc{\char13} variation in respect to the average time given completed tasks \(\Delta_{ii\textsc{\char13}}\) [t.u.] and immediate forbidden sequences, \(Q_{ii\textsc{\char13}}\).}
\label{tab:6}       
\begin{tabular}{p{2.65cm}p{0.9cm}p{0.9cm}p{0.9cm}p{0.9cm}p{0.9cm}p{0.9cm}p{0.9cm}p{0.9cm}}
\hline\noalign{\smallskip}
Task & 1 & 2 & 3 & 4 & 5 & 6 & 7 & 8 \\
\noalign{\smallskip}\hline\noalign{\smallskip}
Task 1 done & \cellcolor[HTML]{EFEFEF} & \cellcolor[HTML]{EFEFEF} & \cellcolor[HTML]{EFEFEF} & \cellcolor[HTML]{EFEFEF} & \cellcolor[HTML]{EFEFEF} & \cellcolor[HTML]{EFEFEF} & 0 & 0 \\
Task 2 done & \cellcolor[HTML]{EFEFEF} & \cellcolor[HTML]{EFEFEF} & -1 & -1.5 & 0 & -1 & 0 & 1 \\
Task 3 done & \cellcolor[HTML]{EFEFEF} & 0 & \cellcolor[HTML]{EFEFEF} & 0 & 0 & 0 & 0 & 0 \\
Task 4 done & \cellcolor[HTML]{EFEFEF} & -0.5 & 0 & \cellcolor[HTML]{EFEFEF} & \cellcolor[HTML]{EFEFEF} & 0 & 0 & 0 \\
Task 5 done & \cellcolor[HTML]{EFEFEF} & -1 & -0.5 & \cellcolor[HTML]{EFEFEF} & \cellcolor[HTML]{EFEFEF} & -2 & 1 & 0 \\
Task 6 done & \cellcolor[HTML]{EFEFEF} & 0 & 0 & 0 & 0 & \cellcolor[HTML]{EFEFEF} & 0 & 0 \\
Task 7 done & 0 & 0 & 0 & 0 & 0 & 0 & \cellcolor[HTML]{EFEFEF} & 0 \\
Task 8 done & 0 & 0 & 0 & 0 & 0 & 0 & 0 & \cellcolor[HTML]{EFEFEF} \\
\noalign{\smallskip}\hline
\end{tabular}
\end{table}

\subsection{Mathematical optimisation solution}
\label{sec:3_3}
To previously assess a baseline solution for the comparison of the Q-learning algorithm performance,  the mathematical optimisation model defined in section 2.2 is implemented in GAMS (29.1.1 ver.), using CPLEX (12.8.0.0 ver.) solver, run in an Intel(R) Core(TM) i7-7700HQ @2.80GHz with 16GB RAM. Given the problem description and data defined in section~\ref{sec:3_2}, the optimal solution obtains an assembly task sequence \(1\rightarrow 8\rightarrow 4\rightarrow 7\rightarrow 5\rightarrow 2\rightarrow 6\rightarrow 3\) with the value of 65 time units (t.u.). The solution complies with the given feasibility constraints for an optimality gap of 0 in under 1 second. The model statistics are presented in Table~\ref{tab:7}. 

\begin{table*}[]
\caption{Mathematical model statistics}
\label{tab:7}       
\begin{tabular}{p{2.2cm}p{1.6cm}p{1.6cm}p{1.9cm}p{1.7cm}p{1.5cm}p{1.4cm}}
\hline\noalign{\smallskip}
\multirow{2}{*}{Model output} & \#Total variables & \#Binary variables & \#Equations & Optimality gap & CPU (s) & Solution \\
\noalign{\smallskip}\cline{2-7}\noalign{\smallskip}
    & 129 & 64 & 550 & 0.0 & 0.34 & 65.0 \\
\noalign{\smallskip}\hline
\end{tabular}
\end{table*}

\subsection{Q-learning algorithm - Scenario I: Learning an assembly sequence based on estimated task average times and variances}
\label{sec:3_4}
To better understand the scenario\textsc{\char13}s results of the Q-learning implementation using Matlab R2020a, it is important to introduce the concept of experiment. An experiment comprises the algorithm\textsc{\char13}s learning phase and the result obtained at the end. The learning phase in an experiment takes place over various episodes dictated by the maximum number of episodes per experiment. In this specific case, an episode starts with no tasks done and ends when all tasks have been successfully completed or when the maximum number of steps has been reached. However, in all scenarios the algorithm considers that, whenever the Q-Learning agent selects an impossible action, the current state does not change, which means that the sequence is penalised for requiring more than 8 steps to complete an episode.

At the end of the experiment, the agent selects the actions based solely on the Q-Values, which means that after learning, the agent always selects the same assembly sequence (expressed as learned assembly sequence or experiment\textsc{\char13}s result). In order to study statistically relevant solutions, each set of parameters and rewards is replicated in 120 experiments.

With the objective of understanding how well the agent would be able to compare the efficiency of the feasible assembly sequences, the rewards are assigned to the processing time of each task. These were then defined through (Eq.~\ref{eq:10}), where \textit{\textbf{R}(s,a)} is the reward for taking the action \textit{a} in the state \textit{s}, \(r_m\) is the reward multiplier, \(r_s\) the reward shift, \(r_p\) the reward penalty and \textit{\textbf{T}(s,a)} the predefined time it takes to complete the action \textit{a} in the state \textit{s}, that is calculated by summing the respective variances to the task\textsc{\char13}s average time:

\begin{equation}\label{eq:10}
\left\{\begin{array}{cl}
R(s,a)=r_m\times (-T(s,a)+r_s) & \;
if\; possible\; action \\ 
R(s,a)=r_p & \;
if\; impossible\; action
\end{array}\right.
\end{equation}

\begin{figure}
  \centering\includegraphics[width=0.85\textwidth]{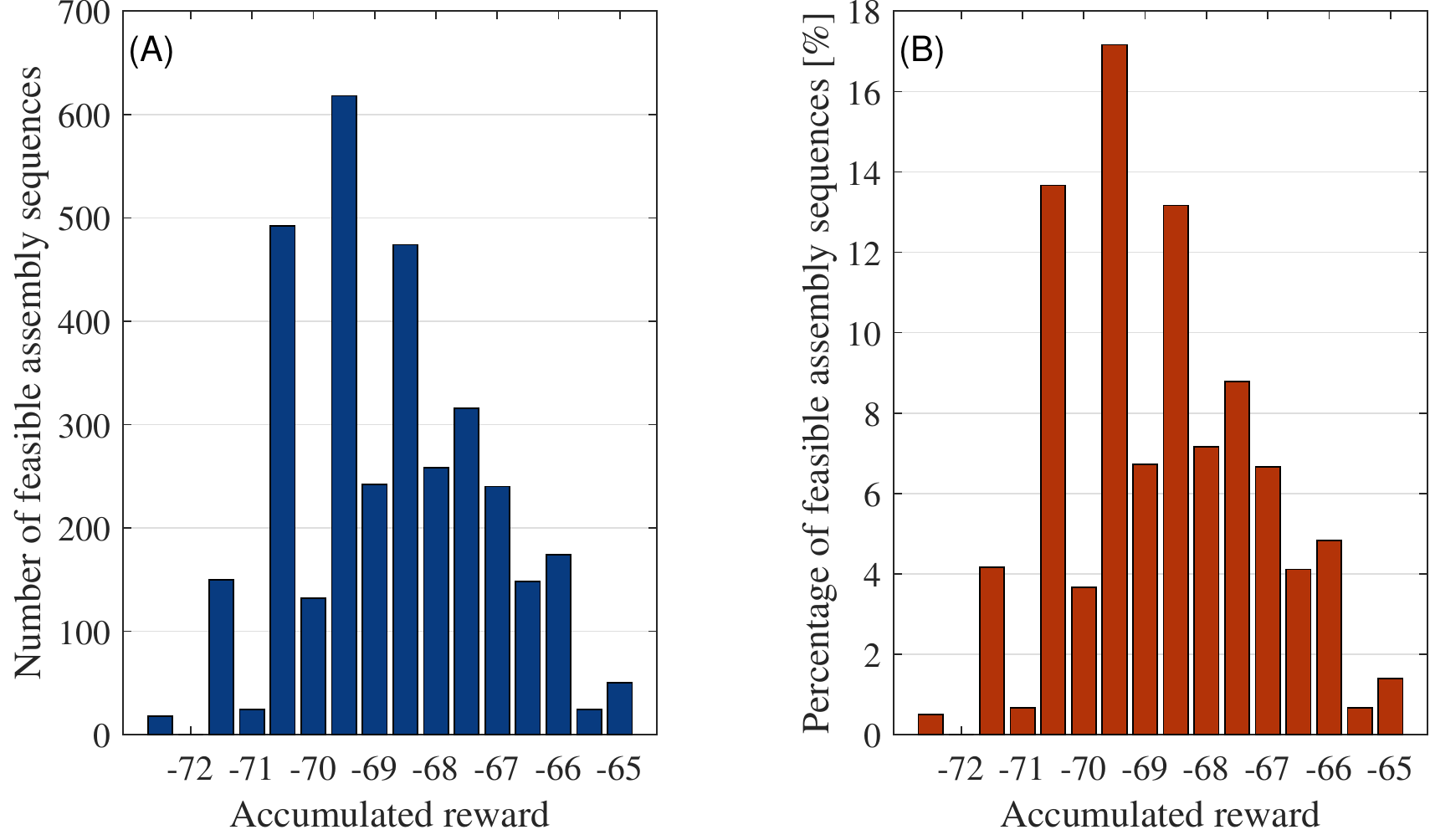}
\caption{Distribution by number (A) and percentage (B) of feasible assembly sequences\textsc{\char13} accumulated rewards.}
\label{fig:5}       
\end{figure}

Since the RL algorithm\textsc{\char13}s goal is to maximise the accumulated reward, in order to learn the most time efficient assembly sequence (minimise the assembly sequence time), the value matrix \textit{\textbf{T}(s,a)} must be subtracted. The \(r_s\) shifts each reward by its value and, as a result, shifts the accumulated reward by eight times its value. The \(r_m\), on the other hand, multiplies the shifted reward. Consequently, the accumulated reward is also multiplied by \(r_m\). If \(r_m\) and \(r_s\) have the values of 1 and 0 respectively, the accumulated reward is equal, in absolute values, to the duration of the assembly sequence.
The accumulated rewards for all feasible assembly sequences for these exact values of \(r_m\) and \(r_s\) are displayed in the Fig.~\ref{fig:5}.

As it can be observed in Figure 5 (B), the most common accumulated reward, corresponding to 18.4\% in all feasible assembly sequences, is -69.5, which correlates to the corresponding total assembly time. An example of such an assembly sequence is \(8\rightarrow 1\rightarrow 3\rightarrow 4\rightarrow 7\rightarrow 2\rightarrow 6\rightarrow 5\) where the accumulated reward is \(\Sigma R(s,a) =-9-10-8-6-11-6.5-7-12=-69.5\). It is also possible to observe in the Fig.~\ref{fig:5} (A) that there are 50 feasible assembly sequences with the maximum accumulated reward of -65 (optimal accumulated reward solution as shown on the mathematical optimization model).

\begin{figure}
  \centering\includegraphics[width=0.99\textwidth]{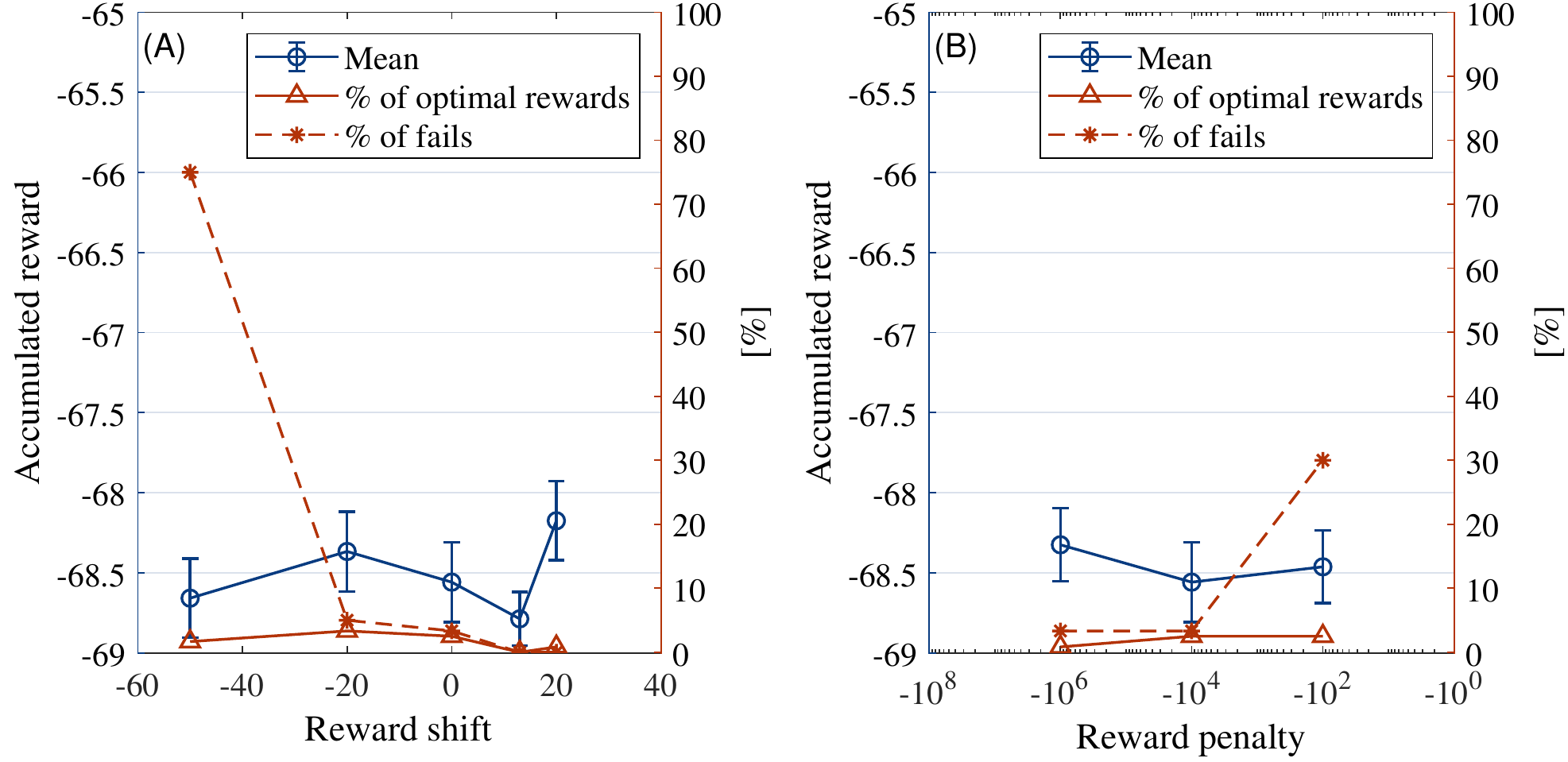}
\caption{Impact of the reward shift (A) and reward penalty (B) on the agent\textsc{\char13}s performance.}
\label{fig:6}       
\end{figure}

In an initial sensitivity analysis, the rewards (\(r_s\) and \(r_p\)) were tested individually (Fig.~\ref{fig:6}) with the parameters in the Table~\ref{tab:6} and with the values 0, 1, and -10000 for the \(r_s\), \(r_m\) and \(r_p\) respectively. The comparison of the performances for the various sets of parameters and rewards considers three indicators: the mean accumulated reward, normalised for a \(r_m\) of 1 and a \(r_s\) of 0; the percentage of times the agent learned one of the 50 optimal assembly sequences; and the percentage of times the agent failed to learn one feasible assembly sequence in the 120 experiments. When the agent failed to learn a feasible assembly sequence, the number of experiments was increased so that the mean would reflect 120 correctly learned assembly sequences. The reason for this rule was the high penalty on the mean of an incorrectly learned assembly, which would turn unfeasible a correct comparison between sets.

When observing the Fig. \ref{fig:6} (A), it can be concluded that a negative \(r_s\) value increases the likelihood of incorrectly learning an impossible assembly sequence because the agent is not able to differentiate between the penalties and the accumulated rewards shifted to values increasingly more negative. Also, a positive \(r_s\) may lead to better results, as seen in the mean increase for the value of 20. The reward penalty, as seen in the Fig~\ref{fig:6} (B), understandably, influences the percentage of fails since an action with a higher \(r_p\) is more likely identified as an incorrect action, i.e. the bigger the penalty the lower the percentage of fails.

With the objective of understanding the impact of the \(r_m\)  and the maximum number of episodes, they were individually changed maintaining the parameters and rewards of the previous sensitivity analysis, except for the \(r_s\) and \(r_p\) with the new values of 20 and -1000000 respectively. Also, in the experiments where the maximum number of episodes was altered, the selected \(r_m\) used was 5.

\begin{figure}
  \centering\includegraphics[width=0.85\textwidth]{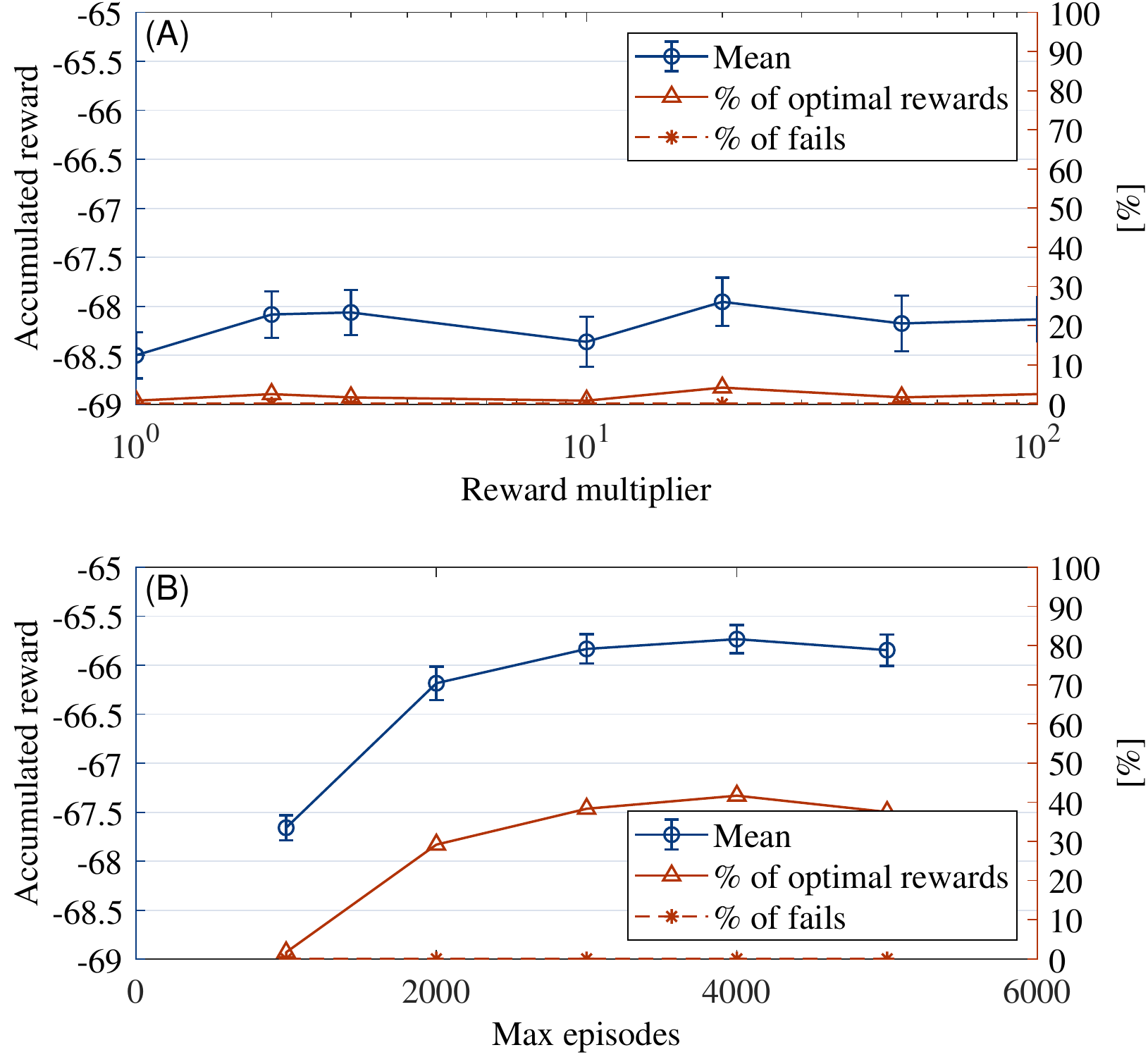}
\caption{Impact of the reward multiplier (A) and maximum number of episodes per experiment (B) on the agent\textsc{\char13}s performance.}
\label{fig:7}       
\end{figure}

In terms of the \(r_m\) sensitivity analysis displayed in the Fig.~\ref{fig:7} (A), there are no statistically significant improvements in the agent\textsc{\char13}s performance. However, the value chosen for \(r_m\) in future sets of parameters and rewards is 20 as it accomplished the best results. In respect to the maximum number of episodes, presented in the Fig~\ref{fig:7} (B), it is possible to conclude that increasing the maximum number of episodes per experiment leads to an increase in performance (both visible in the mean and in the percentage of optimal accumulated rewards) until a certain value in which it seems to plateau. Though, is important to remember that a higher value for the maximum number of episodes is related to the amount of times the experiment has to be repeated for the agent to learn. This means that it is essential to maintain the number as low as possible since in a real scenario it corresponds to the amount of assemblies required for the learning process. Thus, for these parameters, and specially for the epsilon decay of 0.0001, the optimal maximum number of episodes is 3000.

\begin{figure}
  \centering\includegraphics[width=0.85\textwidth]{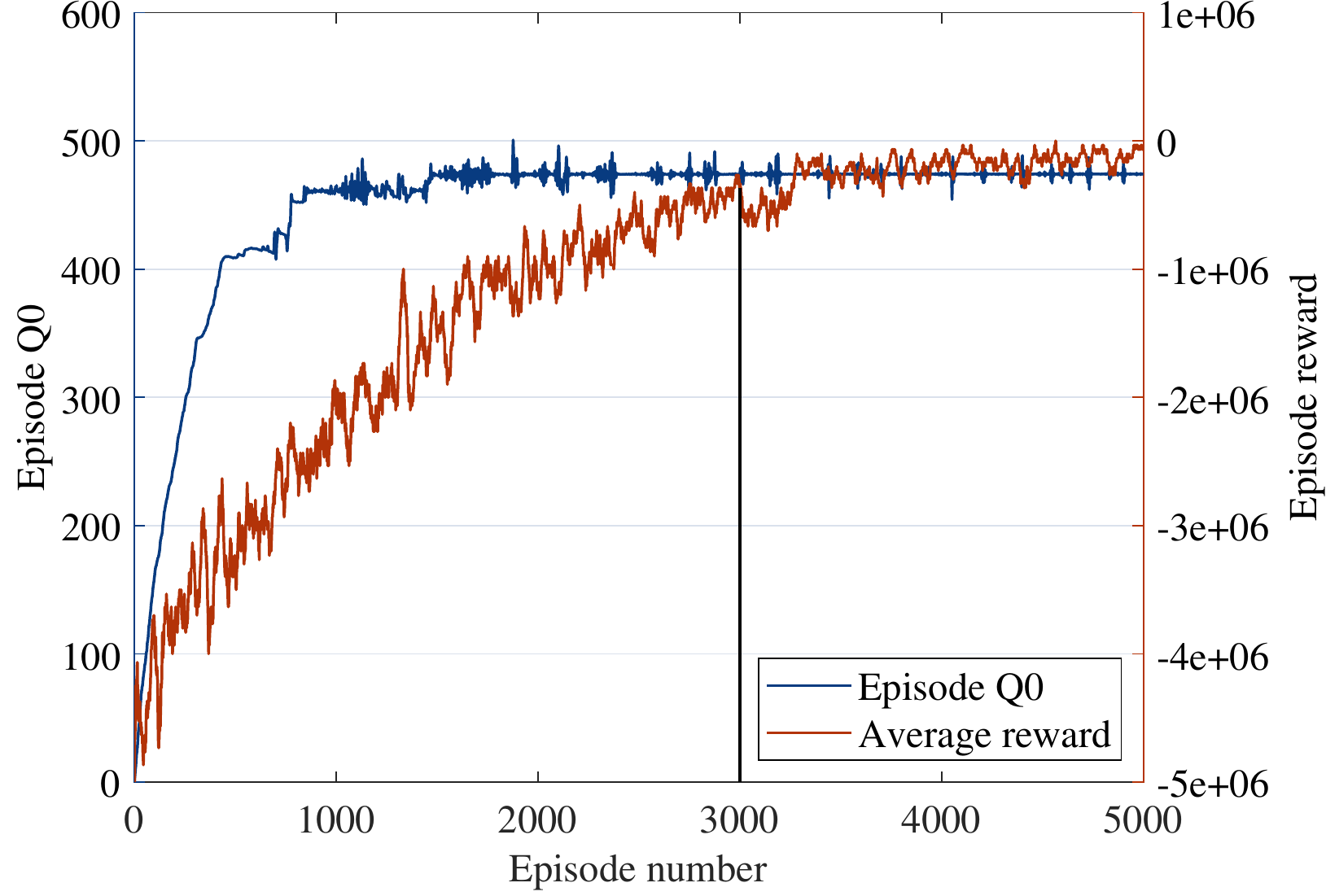}
\caption{Evolution of the episode reward over the episodes for a value of epsilon decay of 0.0001.}
\label{fig:8}       
\end{figure}

The optimal maximum number of episodes is dependent on the epsilon decay\textsc{\char13}s value since a lower epsilon decay leads to a slower increase in the greedy selection by the agent and, as such, requires more episodes in the learning phase. In order to identify the relationship between these two parameters, a graph of the evolution of the episodic accumulated reward in one of the experiments (with 5000 maximum number of episodes) is analysed (Fig.~\ref{fig:8}).

\begin{figure}
  \centering\includegraphics[width=0.94\textwidth]{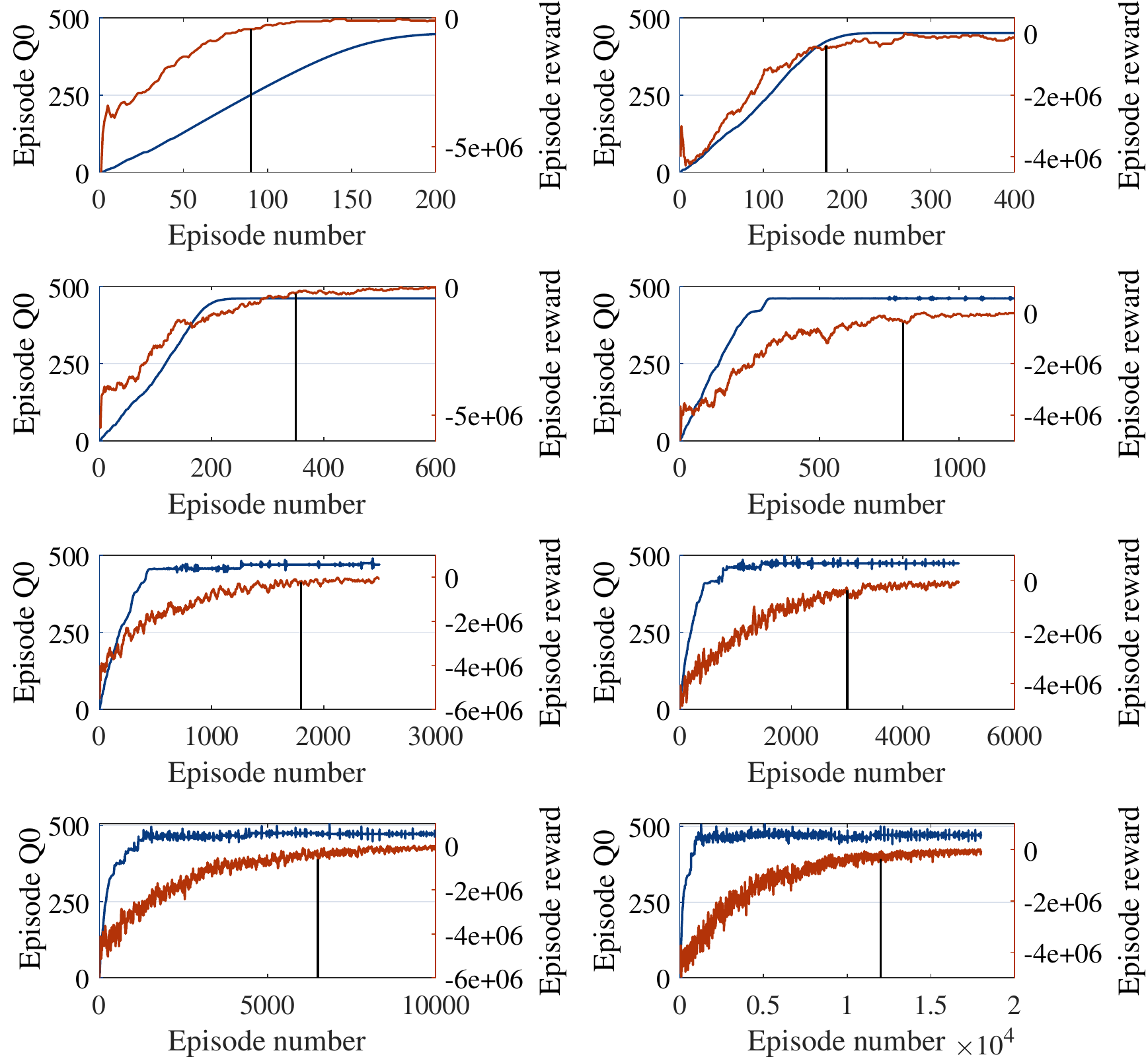}
\caption{Evolution of the episode Q0 (blue line) and episode reward (orange line) over the episodes for various values of epsilon decay.}
\label{fig:9}       
\end{figure}

When analysing the Fig.~\ref{fig:8}, it is possible to observe that after the identified optimal maximum number of episodes the episodic accumulated reward does not greatly increase. This could explain why increasing the maximum number of episodes further does not lead to a significant change in the results in the Fig.~\ref{fig:7}. Similar graphs were analysed for different values of epsilon decay in order to identify in which episode the episodic accumulated plateaus during an experiment to select this value as the optimal maximum number of episodes per experiment for the given epsilon decay (Fig.~\ref{fig:9} and Table~\ref{tab:8}).

\begin{table*}
\caption{Maximum number of episodes selected for each epsilon decay value.}
\label{tab:8}       
\begin{tabular}{p{1.8cm}p{1.0cm}p{1.0cm}p{1.0cm}p{1.0cm}p{1.2cm}p{1.2cm}p{1.30cm}p{1.30cm}}
\hline\noalign{\smallskip}
Epsilon decay & 0.005 & 0.002 & 0.001 & 0.0005 & 0.0002 & 0.0001 & 0.00005 & 0.00003 \\
Max episodes & 90 & 175 & 350 & 800 & 1800 & 3000 & 6500 & 12000 \\
\noalign{\smallskip}\hline
\end{tabular}
\end{table*}

\begin{figure}
  \centering\includegraphics[width=0.95\textwidth]{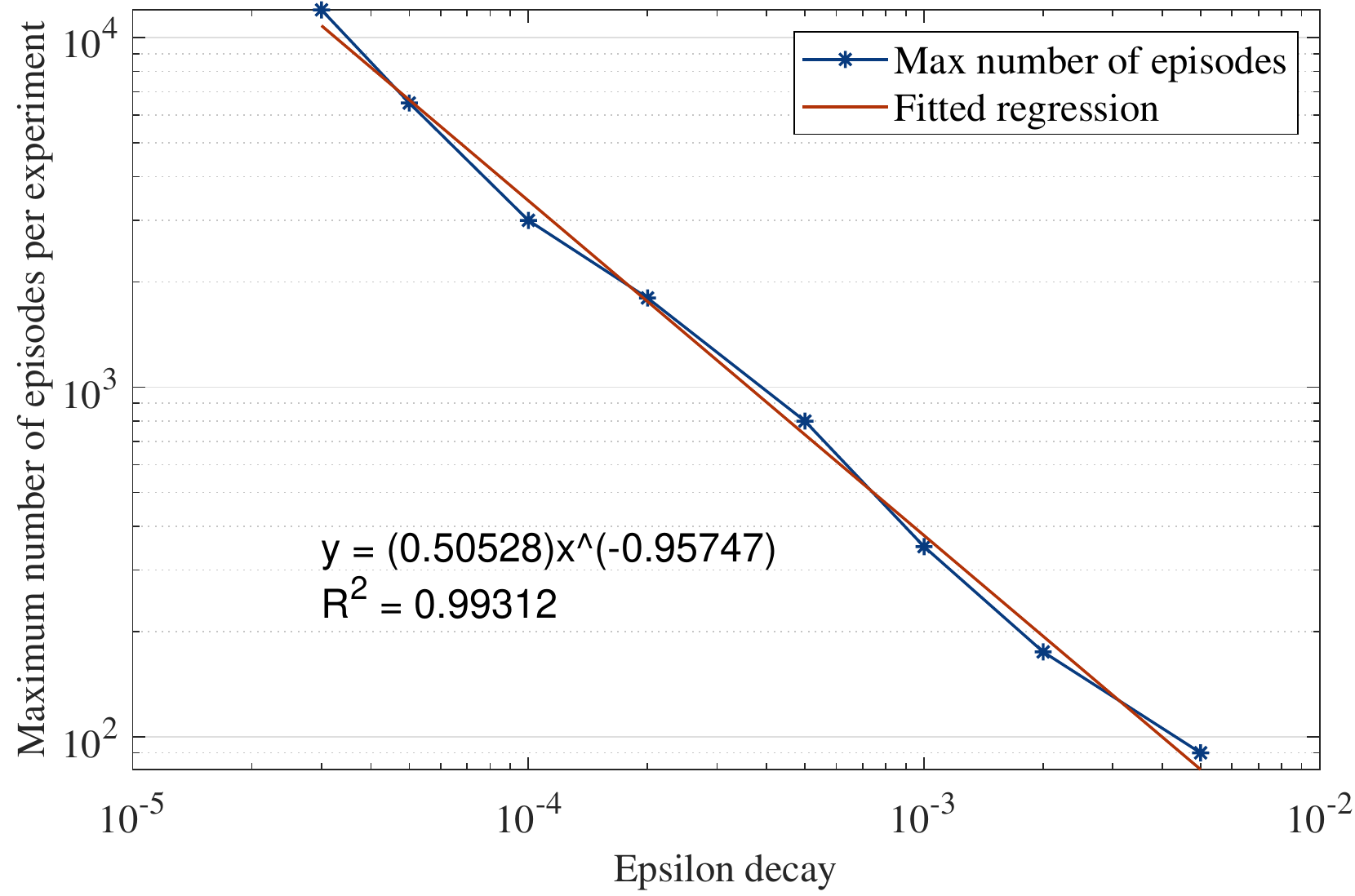}
\caption{Graph of maximum number of episodes over epsilon decay.}
\label{fig:10}       
\end{figure}

From the values available in the Table~\ref{tab:8}, a linear regression was devised using a logarithmic scale in both axes. As shown in the Fig.~\ref{fig:10}, the data approximated fits the function \(y=0.50528\times x^{-0.95747}\) where \textit{y} is the maximum number of episodes and \textit{x} the epsilon decay.

\begin{figure}
  \centering\includegraphics[width=0.95\textwidth]{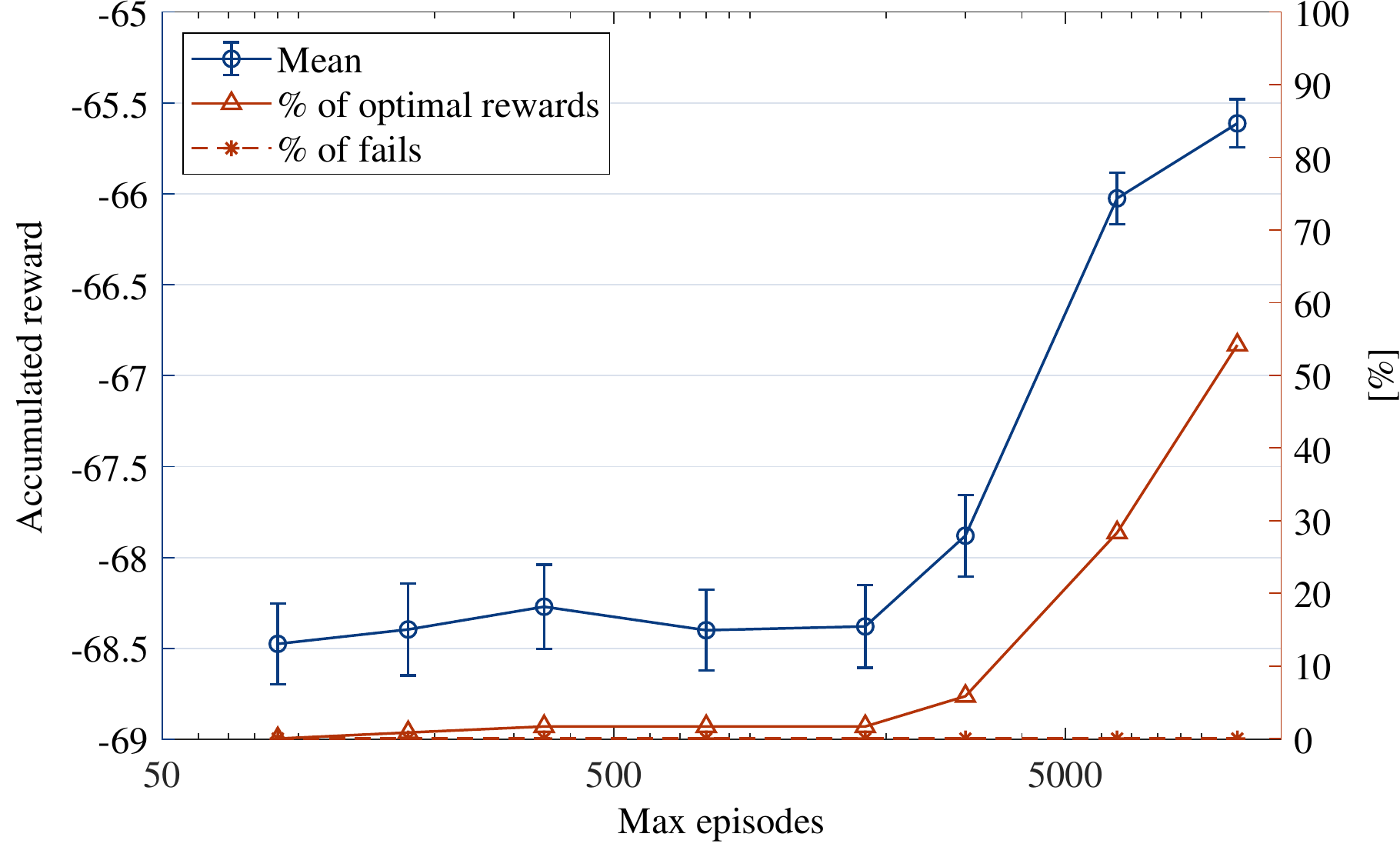}
\caption{Impact of the pairs of epsilon decay and maximum number of episodes on the agent\textsc{\char13}s performance.}
\label{fig:11}       
\end{figure}

Additional new experiments were run with the parameters shown in the Table~\ref{tab:9}, apart from the epsilon decay and maximum number of episodes which were based on the Table~\ref{tab:8}. Results are displayed in the Fig.~\ref{fig:11}.

\begin{table}
\caption{Parameters for the pair epsilon decay and maximum number of episodes experiment.}
\label{tab:9}       
\begin{tabular}{p{7.5cm}p{5cm}}
\hline\noalign{\smallskip}
Parameter & Value \\
\noalign{\smallskip}\hline\noalign{\smallskip}
Learning rate & 1 \\
Discount factor & 1 \\
Epsilon & 0.9 \\
Max steps per episode & 8 \\
Reward shift & 20 \\
Reward multiplier & 20 \\
Reward penalty & -1000000 \\
\noalign{\smallskip}\hline
\end{tabular}
\end{table}

As can be observed in the Fig.~\ref{fig:11}, the increase in the maximum number of episodes, accompanied by the respective decrease in the epsilon decay, leads to better results with an increase in the mean and in the percentage of optimal rewards. It is also possible to notice that such increase only starts around 1800 and 3000 maximum number of episodes.

\subsection{Q-learning algorithm - Scenario II: Learning an assembly sequence based on measured task average times and estimated variances}
\label{sec:3_5}
In the second scenario, task time input data was measured and repeated 10 times (Table~\ref{tab:10}) so that the average processing times are now confirmed (approximation is used for simplification).

\begin{table*}
\caption{Task\textsc{\char13}s time measurements.}
\label{tab:10}       
\begin{tabular}{p{1.8cm}p{0.3cm}p{1cm}p{1cm}p{1cm}p{1cm}p{1cm}p{1cm}p{1cm}p{1cm}}
\hline\noalign{\smallskip}
\multicolumn{2}{l}{Task} & 1 & 2 & 3 & 4 & 5 & 6 & 7 & 8 \\
\noalign{\smallskip}\hline\noalign{\smallskip}
\multirow{10}{*}{\shortstack{Measured\\ tasks\\ {[time units,} \\ t.u.]}} & 1 & 6.36 & 9.10 & 8.59 & 6.75 & 10.62 & 9.87 & 12.36 & 10.06 \\
    & 2 & 6.28 & 9.15 & 10.30 & 8.22 & 10.52 & 11.31 & 12.39 & 8.50 \\
    & 3 & 4.71 & 8.00 & 8.70 & 7.90 & 9.82 & 11.56 & 12.88 & 8.03 \\
    & 4 & 5.38 & 8.15 & 10.49 & 8.29 & 9.25 & 11.00 & 10.49 & 9.09 \\
    & 5 & 5.80 & 8.30 & 8.75 & 7.25 & 9.44 & 8.95 & 11.22 & 8.90 \\
    & 6 & 6.71 & 8.52 & 8.97 & 7.40 & 10.04 & 12.16 & 11.04 & 9.37 \\
    & 7 & 5.73 & 8.17 & 9.00 & 7.95 & 9.07 & 9.24 & 11.94 & 7.70 \\
    & 8 & 7.31 & 8.05 & 8.29 & 6.30 & 10.76 & 10.01 & 10.34 & 8.70 \\
    & 9 & 5.16 & 7.62 & 8.33 & 7.75 & 9.95 & 9.56 & 10.14 & 8.15 \\
    & 10 & 5.89 & 6.55 & 8.50 & 7.00 & 9.30 & 10.09 & 11.98 & 9.71 \\
\noalign{\smallskip}\hline\noalign{\smallskip}
\multicolumn{2}{l}{Mean [t.u.]} & 5.93 & 8.16 & 8.99 & 7.48 & 9.88 & 10.38 & 11.48 & 8.82 \\
\noalign{\smallskip}\hline\noalign{\smallskip}
\multicolumn{2}{l}{Average time} & 6 & 8 & 9 & 7.5 & 10 & 10.5 & 11.5 & 9 \\
\noalign{\smallskip}\hline
\end{tabular}
\end{table*}

As in scenario I, the tasks\textsc{\char13} average times have variations in respect to the corresponding precedence tasks (Table~\ref{tab:11}). It is important to notice that the tasks variability complexity has increased from the scenario I, so that there would be a larger variety of accumulated rewards and a lower number of assembly sequences with the largest accumulated reward, i.e. optimal assembly sequences.

The variability was also increased by introducing the tool changeover time. Since there are two different types of fasteners, the fastening device\textsc{\char13}s tool must be switched during the assembly process. Regarding the fastening device, there are two main assumptions made: the assembly process starts without any tool placed and the tool changeover lasts three time units to be performed. In order to introduce the tool changeover, the number of states must be increased. The states, apart from the binary number that define the completion of each task, can have three possible indexes (0, 1 and 2). The index 0 indicates that the fastening device does not have any tool placed (start of the assembly), the index 1 indicates that the fastening device has the screwdriver applied, and the index 2 indicates the nut driver is applied. The new total number of states is 511, since in the first state the fastening device has no tool and in any other state it can have either the first or the second tool. Due to task dependencies and tool selections, the number of possible states is 149.

\begin{table}[]
\caption{Tasks\textsc{\char13} variation in respect to the average time given completed tasks \(\Delta_{ii\textsc{\char13}}\) [t.u.] and immediate forbidden sequences, \(Q_{ii\textsc{\char13}}\).}
\label{tab:11}       
\begin{tabular}{p{2.65cm}p{0.9cm}p{0.9cm}p{0.9cm}p{0.9cm}p{0.9cm}p{0.9cm}p{0.9cm}p{0.9cm}}
\hline\noalign{\smallskip}
Task & 1 & 2 & 3 & 4 & 5 & 6 & 7 & 8 \\
\noalign{\smallskip}\hline\noalign{\smallskip}
Task 1 done & \cellcolor[HTML]{EFEFEF} & \cellcolor[HTML]{EFEFEF} & \cellcolor[HTML]{EFEFEF} & \cellcolor[HTML]{EFEFEF} & \cellcolor[HTML]{EFEFEF} & \cellcolor[HTML]{EFEFEF} & 0 & 0 \\
Task 2 done & \cellcolor[HTML]{EFEFEF} & \cellcolor[HTML]{EFEFEF} & -2 & -3 & -0.5 & -2 & 0 & 1.5 \\
Task 3 done & \cellcolor[HTML]{EFEFEF} & 0 & \cellcolor[HTML]{EFEFEF} & 0 & 0 & 0 & 0 & 0 \\
Task 4 done & \cellcolor[HTML]{EFEFEF} & -1 & 0 & \cellcolor[HTML]{EFEFEF} & \cellcolor[HTML]{EFEFEF} & -1.5 & 0 & 0 \\
Task 5 done & \cellcolor[HTML]{EFEFEF} & -2 & -1 & \cellcolor[HTML]{EFEFEF} & \cellcolor[HTML]{EFEFEF} & -3 & 2 & 0 \\
Task 6 done & \cellcolor[HTML]{EFEFEF} & -1 & -0.5 & 1 & -0.5 & \cellcolor[HTML]{EFEFEF} & 0 & 0 \\
Task 7 done & 0 & 0 & 0 & 0 & 0 & 0 & \cellcolor[HTML]{EFEFEF} & 0 \\
Task 8 done & 0 & 0 & 0 & 0 & 0 & 0 & 0 & \cellcolor[HTML]{EFEFEF} \\
\noalign{\smallskip}\hline
\end{tabular}
\end{table}

\begin{figure}
  \centering\includegraphics[width=0.87\textwidth]{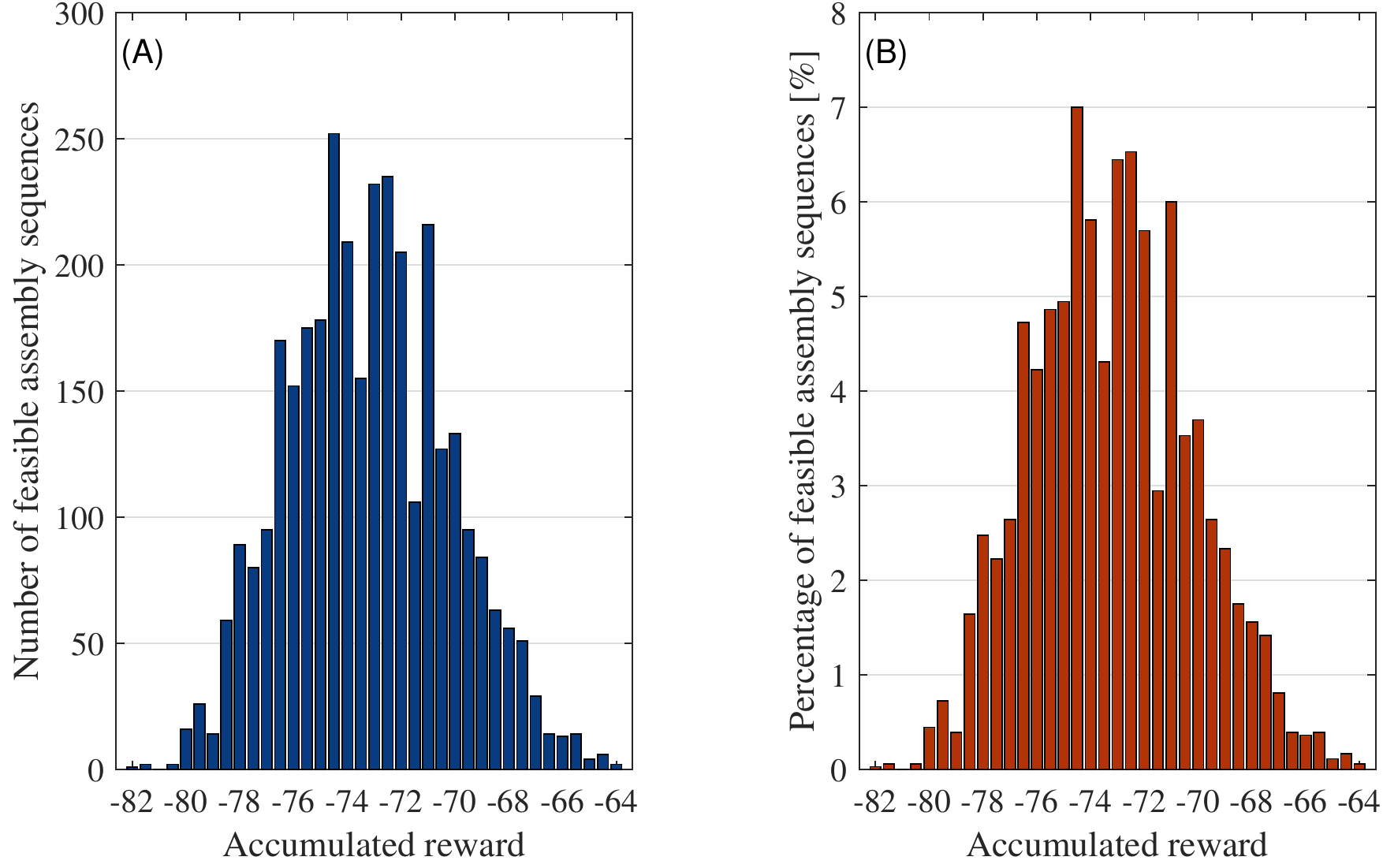}
\caption{Distribution by number (A) and percentage (B) of feasible assembly sequences\textsc{\char13} accumulated rewards.}
\label{fig:12}       
\end{figure}

With the respective changes to the average times and time variances the new distribution of accumulated rewards from the feasible assembly sequences can be observed in the Fig.~\ref{fig:12}.

\begin{figure}
  \centering\includegraphics[width=0.87\textwidth]{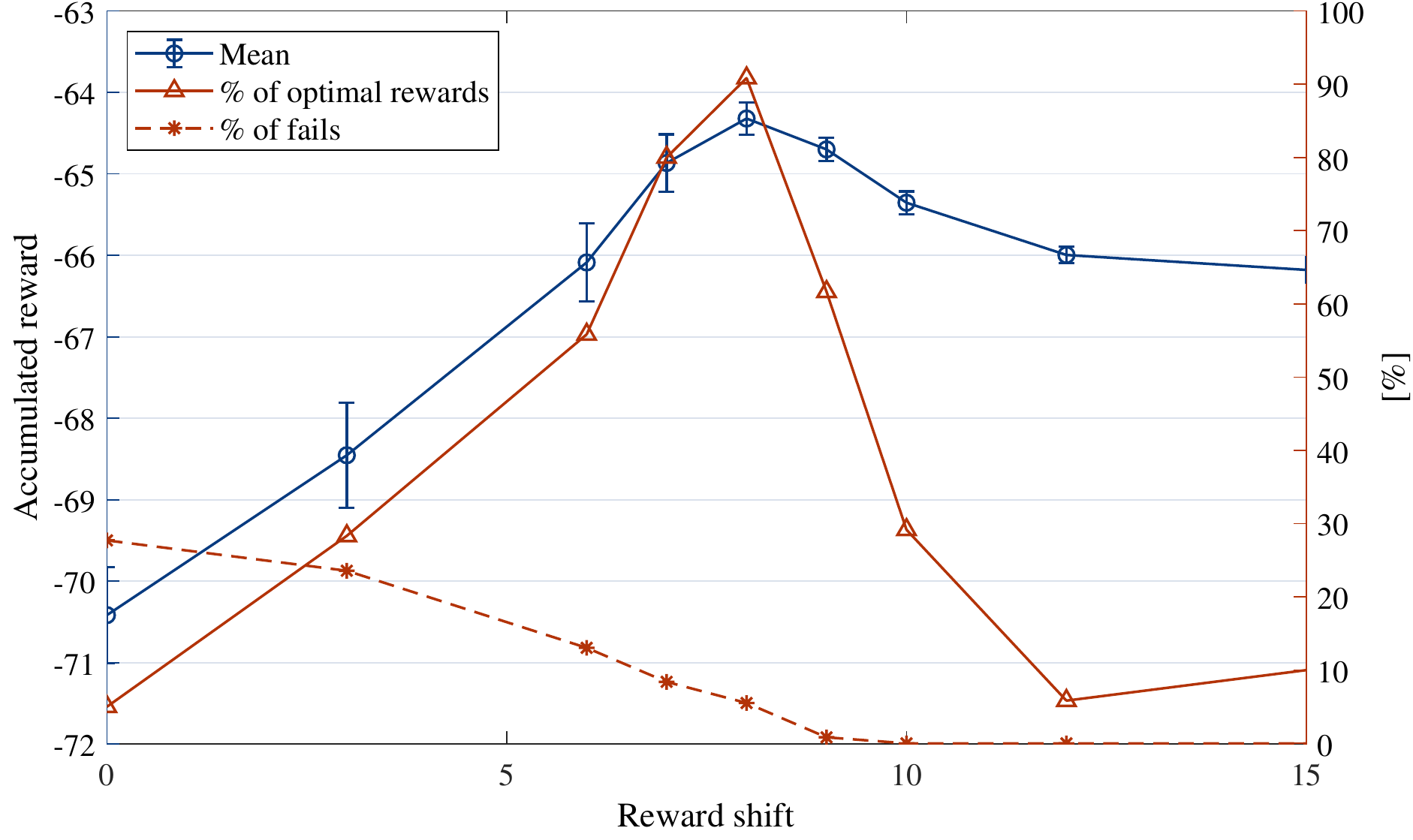}
\caption{Reward shift\textsc{\char13}s impact on the agent\textsc{\char13}s performance.}
\label{fig:13}       
\end{figure}

The new distribution has, as previously stated, a larger variance of accumulated rewards and the highest accumulated reward or optimal accumulated reward is shared only by 2 assembly sequences (Fig.~\ref{fig:12} (B)), which are \(7\rightarrow 1\rightarrow 8\rightarrow 2\rightarrow 4\rightarrow 5\rightarrow 6\rightarrow 3\) and \(7\rightarrow 8\rightarrow 1\rightarrow 2\rightarrow 4\rightarrow 5\rightarrow 6\rightarrow 3\), and has the value of -64. In this new scenario, it may be easier to understand the impact of the set\textsc{\char13}s parameters on the agent\textsc{\char13}s performance. The reward shift was modified for the values (0, 3, 6, 7, 8, 9, 10 12, 15) while using the values of the Table~\ref{tab:9}, apart from the epsilon decay and maximum number of episodes, which had the values of 0.00005 and 6500 respectively. The results of the multiple sets of 120 experiments are displayed in the Fig.~\ref{fig:13}.

\begin{figure}
  \centering\includegraphics[width=0.87\textwidth]{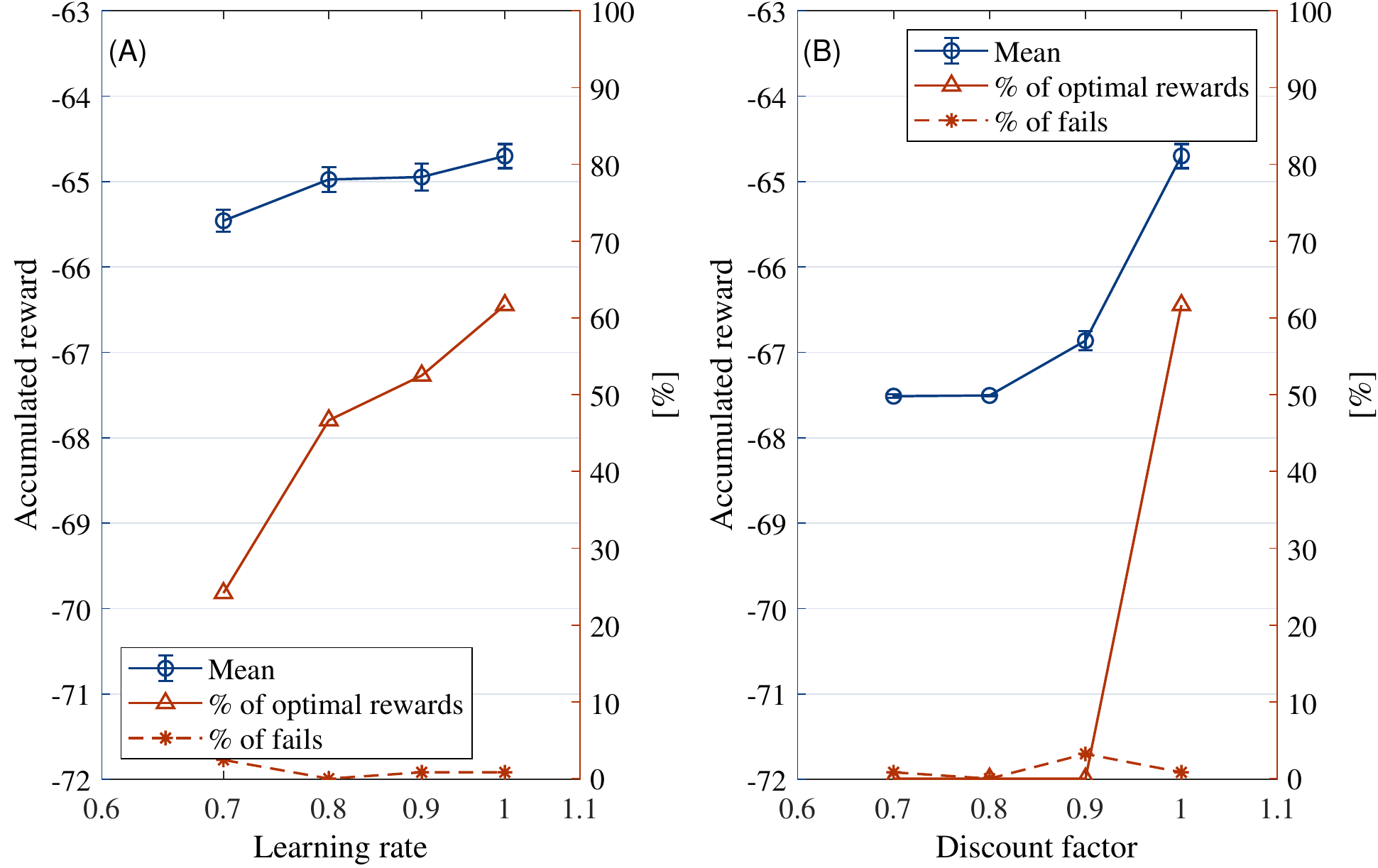}
\caption{Learning rate\textsc{\char13}s and discount factor\textsc{\char13}s impact on the agent\textsc{\char13}s performance.}
\label{fig:14}       
\end{figure}

The mean of all the 37 possible values of accumulated reward is -73 (for a \(r_s\) of 0 and \(r_m\) of 1), and therefore, if subdivided evenly, each task would have a reward of -9.125, which we will define as mean task reward. A value of \(r_s\) equal to the mean task reward shifts the accumulated rewards to a position where they are evenly separated into positive and negative. When analysing the Fig.~\ref{fig:13} it is possible to identify that the best accumulated reward occurs for a value of the reward shift of 8. Also, it is important to notice that the percentage of fails decreases with the increase of the reward shift and is approximately 0 for values higher or equal to 9. Thus, the optimal reward shift may be related to the mean task reward, but it may be relevant to confirm this relationship with a different scenario. A value of the reward shift slightly lower than the mean task reward may improve the mean accumulated reward since a larger number of the accumulated rewards are negative. However, it also increases the percentage of fails, which is highly prejudicial for a real scenario. For that reason, the optimal value for the reward shift is 9. With this new \(r_s\) value, the learning rate and the discount factor were individually analysed (Fig.~\ref{fig:14}).

\begin{figure}
  \centering\includegraphics[width=0.87\textwidth]{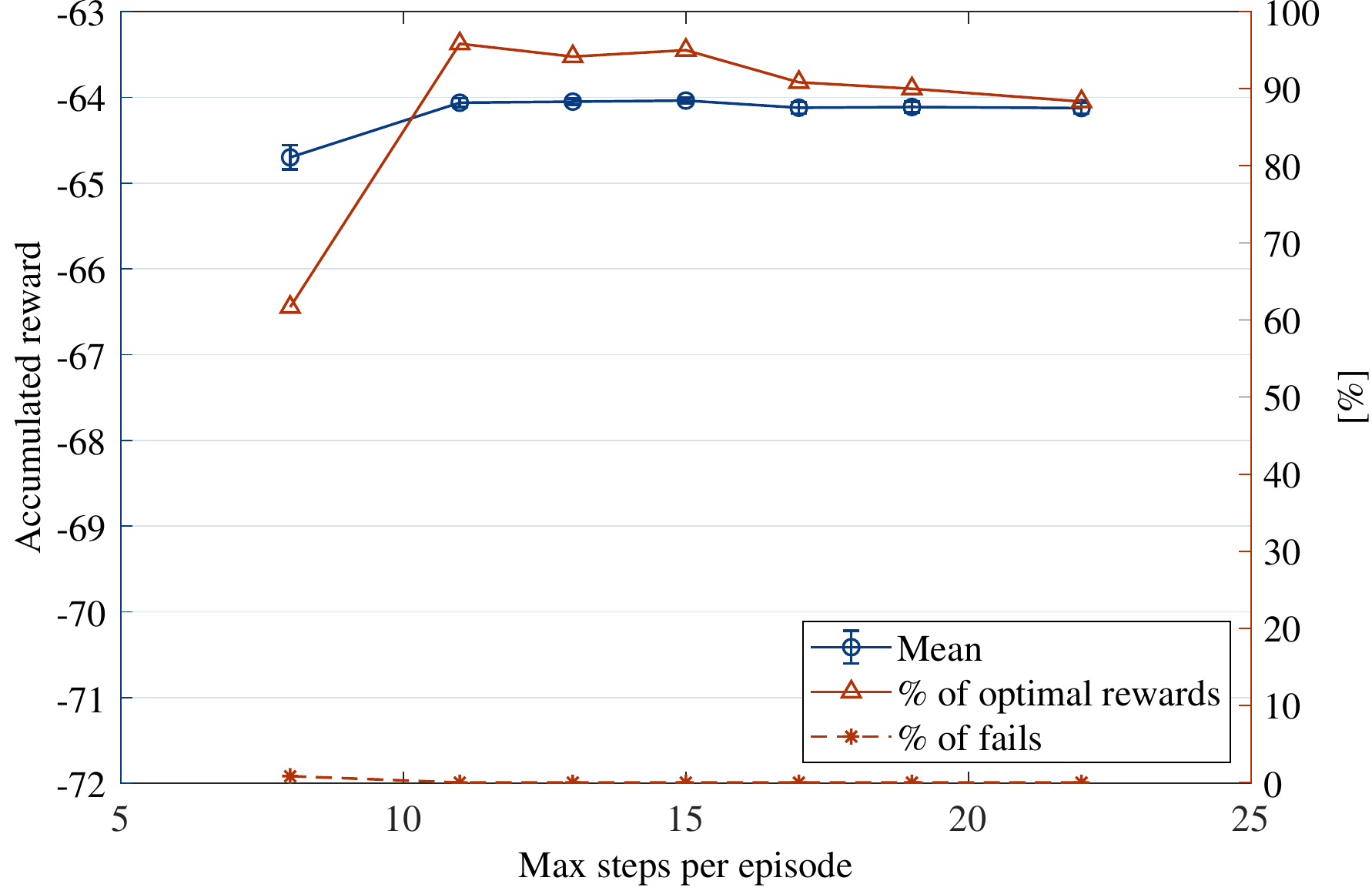}
\caption{Maximum steps per episode\textsc{\char13}s impact on the agent\textsc{\char13}s performance.}
\label{fig:15}       
\end{figure}

Since in both cases the confidence interval is very small, even small differences in the mean are significant. Both in the learning rate and in the discount factor, it can be concluded that the optimal value is 1, however, in the discount factor case the difference in the mean and in the percentage of optimal results is more accentuated. Then, with the same set\textsc{\char13}s parameters as before, the maximum steps per episode were individually analysed (Fig.~\ref{fig:15}).

It is possible to conclude that an early increase in the maximum number of steps per episode leads to a significant increase both in the mean and in the percentage of optimal rewards. This shows that additional steps beyond the number of assembly tasks slightly contributes to improve the agent learning process, although further increase in this value does not significantly alter the results. The value 15 was selected for this parameter.

\begin{figure}
  \centering\includegraphics[width=0.87\textwidth]{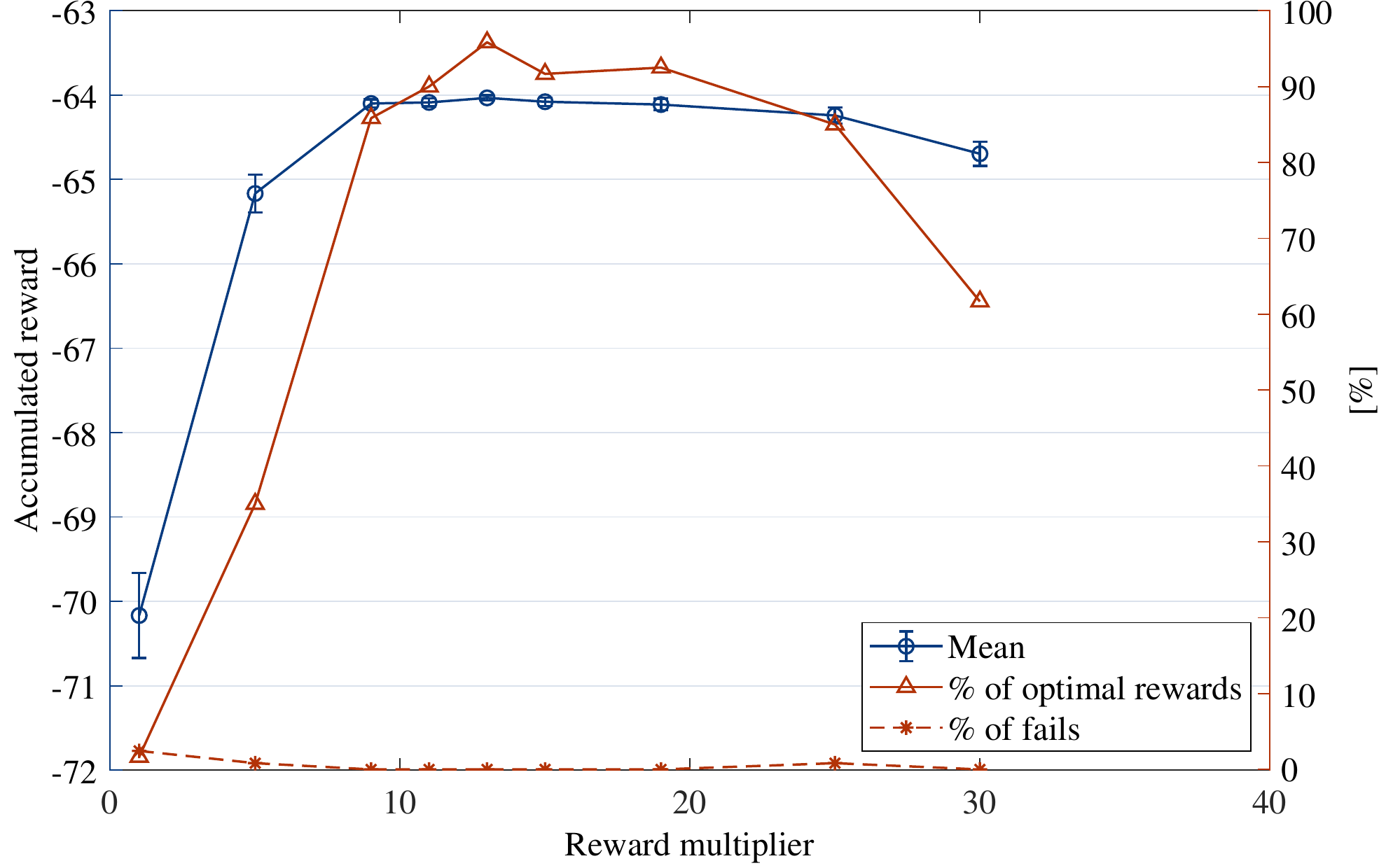}
\caption{Reward multiplier\textsc{\char13}s impact on the agent\textsc{\char13}s performance.}
\label{fig:16}       
\end{figure}

Lastly, with all the other parameters decided, the reward multiplier\textsc{\char13}s impact was studied with various values (1, 5, 9, 11, 13, 15, 19, 25, 30) and the experiments results are visible in the Fig.~\ref{fig:16}.

\begin{figure}
  \centering\includegraphics[width=0.87\textwidth]{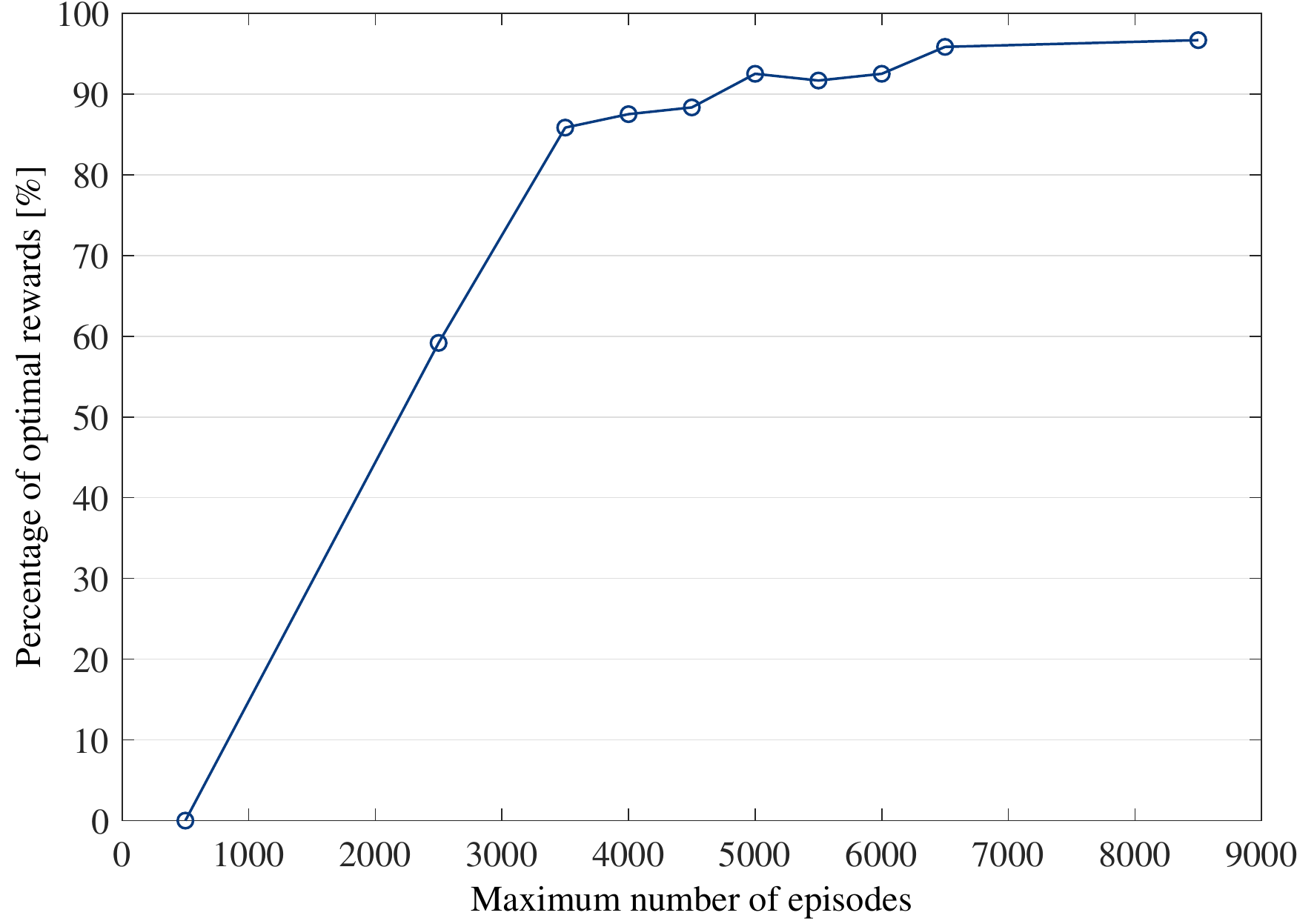}
\caption{Evolution of the percentage of optimal rewards in regards to the maximum number of episodes.}
\label{fig:17}       
\end{figure}

It can be observed that there is an optimal value for \(r_m\), since the increase in the value of the reward multiplier leads to a steady increase both in the mean and in the percentage of optimal rewards. For the smaller values of the reward multiplier, the percentage of fails is nonzero. Based on the graph, it can be defined that the optimal reward multiplier value is 13.
To confirm the choice of 6500 maximum number of episodes, in the Fig.~\ref{fig:17} are plotted the agent\textsc{\char13}s performances for various values of maximum number of episodes.

As expected, a reduction in the maximum number of episodes leads to a lower percentage of optimal rewards. Therefore, in order to guarantee the threshold of 95\% the value of the maximum number of episodes is maintained.
After the previous scenarios experiments, it is possible to conclude that the best set\textsc{\char13}s parameters are the ones expressed in the Table~\ref{tab:12}, with which the agent, in 120 experiments with 6500 episodes, was able to learn one of the 2 optimal assembly sequences 115 times \((\approx 95.8\%)\) (Fig.~\ref{fig:17}), one of the assembly sequences with the second best accumulated reward 4 times \((\approx 3.3\%)\) and one of the fifth best accumulated reward once \((\approx 0.8\%)\), while never failing to learn a feasible assembly sequence.

\begin{table}
\caption{Optimal set\textsc{\char13}s parameters.}
\label{tab:12}       
\begin{tabular}{p{7.5cm}p{5cm}}
\hline\noalign{\smallskip}
Parameter & Value \\
\noalign{\smallskip}\hline\noalign{\smallskip}
Learning rate & 1 \\
Discount factor & 1 \\
Epsilon & 0.9 \\
Epsilon decay & 0.00005 \\
Max steps per episodes & 15 \\
Max episodes & 6500 \\
Reward shift & 9 \\
Reward multiplier & 13 \\
Reward penalty & -1000000 \\
\noalign{\smallskip}\hline
\end{tabular}
\end{table}

\subsection{Q-learning algorithm - Scenario III: Learning an assembly sequence based on measured task average times and estimated variances with restricted actions}
\label{sec:3_6}
\begin{figure}
  \centering\includegraphics[width=0.87\textwidth]{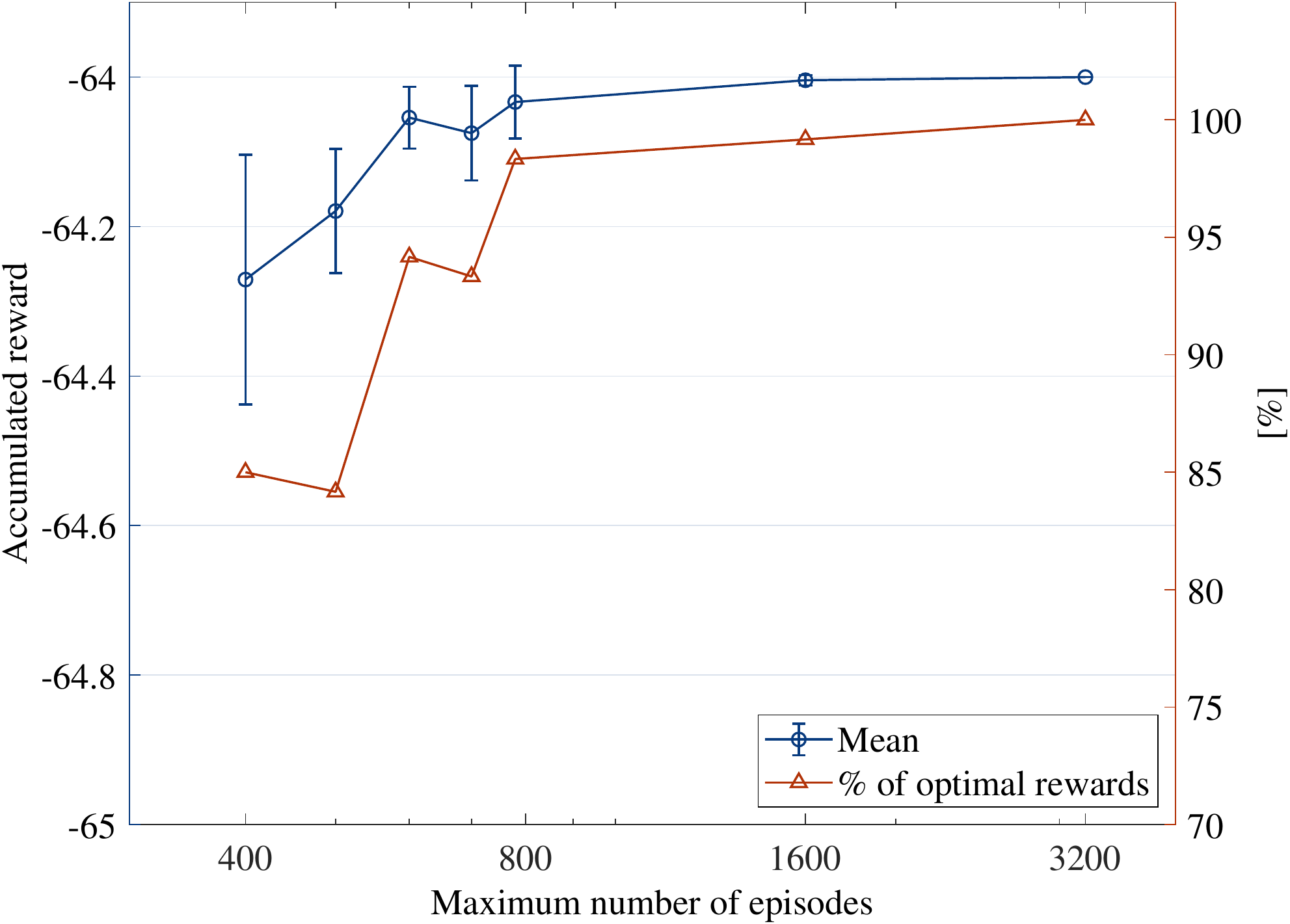}
\caption{Evolution of the mean accumulated reward and the percentage of optimal rewards in regards to the maximum number of episodes.}
\label{fig:18}       
\end{figure}

As previously discussed, the agent is capable of selecting impossible actions, which are penalised. However, this greatly increases the exploration required to achieve a correct assembly sequence. Therefore, in this final scenario, impossible actions were restricted, instead of being penalised in order to identify the possibility of reducing the number of episodes. Apart from the maximum number of episodes, which had the values of (400, 500, 600, 700, 750, 780, 1600, 3200), the epsilon decay, which had the corresponding values calculated from the regression equation in the Fig.~\ref{fig:10} and the reward penalty that is now nonexistent the parameters used in the experiments are displayed in Table~\ref{tab:12}, Fig.~\ref{fig:18}.

It can be observed that by restricting impossible actions the maximum number of episodes can be reduced to 780 episodes, which corresponds to only 21.7\% of all possible assembly sequences, while maintaining a percentage of optimal rewards of 98.3\%.

\section{Conclusions}
\label{sec:4}

In this work, the challenges in the application of a reinforcement learning algorithm in a assembly sequence problem are studied, considering the implementation of a Q-Learning model-free algorithm. By formulating the problem as a MDP, it is shown that Q-Learning finds an optimal state-action policy that maximises the accumulated reward over a succession of given steps. This allows to verify the application of a scalable method to address the optimisation of an assembly sequence of an object as a sequential decision process, where the action in one state influences the transition to the subsequent state. Despite its recent application in the literature, RL methods show a straightforward versatility in complex problems where uncertainty plays a significant role, such as on-line industrial environments, where until now mostly traditional exact and non-exact optimization approaches are considered.
The improvement efficiency of the assembly time of complex products is often impractical due to the complexity of assessing all tasks combinations. This approach has the advantage of achieving suitable optimisation results, where the optimal assembly sequence was learned 98.3\% of the times. To guarantee this threshold (over 95\%), the algorithm requires 780 assemblies (episodes) to correctly learn the best assembly sequence, which is corresponds to approximately 21.7\% of the number of feasible assembly sequences. It is acknowledged that an increasingly complex assembly processes might reveal an unpractical use of Q-Learning algorithm due to the “curse of dimensionality”.
In future work, the acquisition of the tasks\textsc{\char13} durations on-line during real assemblies will be considered.

\section{Conflict of interest}
The authors declare that there is no conflict of interest. 

\section{Declaration of Competing Interest}
The authors report no declarations of interest. 

\section{Funding}
This research was partially supported by project PRODUTECH4S\&C (46102) by UE/FEDER through the program COMPETE 2020 and the Portuguese Foundation for Science and Technology (FCT): COBOTIS (PTDC/EME-EME/32595/2017) and UIDB/00285/2020.






\bibliographystyle{elsarticle-num-names}







\end{document}